\documentclass[10pt,journal,compsoc]{IEEEtran}

\usepackage{amsmath}
\usepackage{amssymb}
\usepackage{booktabs}
\usepackage{epsfig}
\usepackage{enumitem}
\usepackage{multirow}
\usepackage{colortbl}
\usepackage{setspace}
\usepackage{bigstrut}
\usepackage{microtype}      
\usepackage[american]{babel}
\usepackage[caption=false]{subfig}
\usepackage{url}
\usepackage{bm}
\usepackage{color}
\newcommand{\eg}{\emph{e.g.}}

\newcommand{\ie}{\emph{i.e.}}

\graphicspath{{figure/}}

\ifCLASSOPTIONcompsoc
  \usepackage[nocompress]{cite}
\else
  \usepackage{cite}
\fi

\ifCLASSINFOpdf
\else
\fi

\begin{document}
%

\title{Delving Deep into Simplicity Bias for Long-Tailed Image Recognition}

\author{
	Xiu-Shen Wei,~\IEEEmembership{Member,~IEEE}, Xuhao Sun, Yang Shen, Anqi Xu, Peng Wang,~\IEEEmembership{Member,~IEEE}, Faen Zhang
\IEEEcompsocitemizethanks{\IEEEcompsocthanksitem {X.-S. Wei, X. Sun and Y. Shen are with School of Computer Science and Engineering, Nanjing University of Science and Technology, China. A. Xu is with the University of Toronto. P. Wang is with School of Computing and Information Technology, University of Wollongong, Australia. F. Zhang is with Qingdao AInnovation Technology Group Co., Ltd.}
\IEEEcompsocthanksitem{The first two authors contributed equally to this work. P. Wang is the corresponding author.}}
}

\markboth{SUBMITTED TO IEEE TPAMI}%
{Wei \MakeLowercase{\textit{et al.}}: Delving Deep into Simplicity Bias for Long-Tailed Image Recognition}
%

\IEEEtitleabstractindextext{%
\begin{abstract}
Simplicity Bias (SB) is a phenomenon that deep neural networks tend to rely favorably on simpler predictive patterns but ignore some complex features when applied to supervised discriminative tasks. In this work, we investigate SB in long-tailed image recognition and find the tail classes suffer more severely from SB, which harms the generalization performance of such underrepresented classes. We empirically report that self-supervised learning (SSL) can mitigate SB and perform in complementary to the supervised counterpart by enriching the features extracted from tail samples and consequently taking better advantage of such rare samples. However, standard SSL methods are designed without explicitly considering the inherent data distribution in terms of classes and may not be optimal for long-tailed distributed data. To address this limitation, we propose a novel SSL method tailored to imbalanced data. It leverages SSL by triple diverse levels, \ie, holistic-, partial-, and augmented-level, to enhance the learning of predictive complex patterns, which provides the potential to overcome the severe SB on tail data. Both quantitative and qualitative experimental results on five long-tailed benchmark datasets show our method can effectively mitigate SB and significantly outperform the competing state-of-the-arts.
\end{abstract}

\begin{IEEEkeywords}
Long-Tailed Image Recognition; Simplicity Bias; Deep Learning; Self-Supervised Learning.
\end{IEEEkeywords}}

\maketitle


\IEEEdisplaynontitleabstractindextext

%
\IEEEpeerreviewmaketitle

\section{Introduction}

\IEEEPARstart{D}{eep} neural networks have revolutionized various computer vision applications such as image recognition, object detection, and semantic segmentation, just name a few. However, even the most powerful deep model suffers when learning from long-tailed distributed data. The major bottleneck lies on the unsatisfactory performance of the tail classes which are underrepresented by limited samples. Thus, at the core of long-tailed image recognition lies the strategies to mitigating such data imbalance, where common solutions include re-sampling~\cite{he2009learning, buda2018systematic}, loss re-weighting~\cite{huang2019deep, cb-focal, ldam, khan2019striking}, and data augmentation~\cite{feataugeccv20}, etc.

In this work, we inspect long-tailed image recognition through the lens of simplicity bias (SB)~\cite{simpBiasNIPs20, OOD, huh2021low}. SB is a phenomenon that when applied to supervised discriminative tasks such as image recognition, deep neural networks tend to rely favorably on simpler predictive patterns but ignore some complex features that can be important for model generalization. To unveil if there exists correlation between SB and the amount of data available for a specific class, we investigate SB in long-tailed image recognition and uncover that, under standard supervised losses such as cross-entropy, tail classes suffer more severely from SB. As shown in  Figure~\ref{fig:pre_exp}, for underrepresented classes, simple patterns such as the ``digits'' are activated but complex patterns such as the ``truck'' are ignored. This observation partially explains the poor generalization performance on tail classes.

On the other side, different from supervised learning that aims to identify image patterns that are sufficient to distinguish the limited training samples of each particular class, self-supervised learning (SSL) is expected to encourage more comprehensive features to learn augmentation-invariant image representations~\cite{gidaris2019boosting}. Motivated by this, in our preliminary study we combine off-the-shelf SSL and supervised training together on long-tailed image recognition and empirically report that the SSL component can mitigate SB and performs in complementary to the supervised counterpart by enriching the features from tail samples and consequently taking better advantage of such rare samples. However, standard SSL methods are designed without explicitly considering the inherent distribution of the data in terms of classes and thus may not function optimally for long-tailed distributed data. To address such limitation, we propose a novel SSL method termed as ``Triple-Level Self-Supervised Learning'' (3LSSL) which is tailored to imbalanced data. More specifically, 3LSSL first performs the holistic-level SSL to enhance the learning of comprehensive patterns in the holistic level w.r.t. the raw input. It then develops a partial-level SSL by leveraging a masking manner to force the models to learn more comprehensive (complementary) information from the partial level w.r.t. the holistic level. To further consider the severe SB on tail data, it provides pseudo positive samples for an anchor sample based on the prediction results from the classifier, which essentially plays the role of augmenting the tail classes and forms as the augmented-level SSL.

To evaluate our method, we conduct extensive experiments on five benchmark long-tailed recognition datasets, \ie, \emph{long-tailed CIFAR-10}/\emph{long-tailed CIFAR-100}~\cite{cifar}, \emph{ImageNet-LT}~\cite{oltr}, \emph{Places-LT}~\cite{zhou2017places} and \emph{iNaturalist 2018}~\cite{van2018inaturalist}. Quantitative results of classification accuracy on these datasets show that our 3LSSL method consistently surpasses existing state-of-the-art methods by a large margin. Also, the ablation studies of these crucial components in our method also validate their own effectiveness. In particular, to justify the effects of alleviation of simplicity bias, we demonstrate various qualitative results for visualization analyses. Beyond that, we expect our work can provide new understanding about learning from long-tailed data and inspire new ideas to address SB in long-tailed or rare data.  

The rest of the paper is organized as follows. Section~\ref{sec:related} retrospects the related work. Section~\ref{sec:SB} elaborates our investigation about simplicity bias in long-tailed image recognition tasks. Section~\ref{sec:approach} introduces the details of our proposed method. Experiments and analyses of both qualitative and quantitative aspects are provided in Section~\ref{sec:experiments}. Section~\ref{sec:conc} presents conclusions and future work.

\section{Related Work}\label{sec:related}

We briefly review the related work in the following three aspects, \ie, long-tailed image recognition, simplicity bias in neural networks, and self-supervised learning.

\subsection{Long-Tailed Image Recognition}

Long-tailed image recognition is a fundamental research topic in machine learning and computer vision, where it aims to overcome the data imbalance challenge~\cite{zhihuatkde,he2009learning,cb-focal}. In general, existing long-tailed image recognition methods can be roughly separated into the following paradigms. 1) Class re-balancing strategies: Re-balancing strategies, \eg, data re-sampling~\cite{shen2016relay,japkowicz2002class} and loss re-weighting~\cite{zhihuaicdm,huang2016learning}, are conventional solutions for handling long-tailed distributed data. The popular re-sampling methods involve over-sampling by simply repeating data of minority classes~\cite{shen2016relay, buda2018systematic,  byrd2019effect} and under-sampling by abandoning data of dominant classes~\cite{japkowicz2002class,buda2018systematic,he2009learning}. But with re-sampling, duplicated tailed samples might lead to over-fitting upon minority classes~\cite{chawla2002smote,cb-focal}, while discarding precious data will inevitably impair the generalization ability of deep networks. On the other hand, {re-weighting} belongs to another line of class re-balancing strategies, which works by allocating large weights for training samples of tail classes in loss functions, \eg,~\cite{huang2016learning, wang2017learning}. 2) Decoupled learning: It is a recent trend towards effective long-tailed image recognition, which decouples the image representation learning and classifier learning to improve long-tailed classification performance. Specifically, Zhou et al.~\cite{bbn} and Kang et al.~\cite{decouple} decoupled classifier and representation learning and found that uniform sampling could benefit representation learning while classifier learning favors class-balance sampling. Later, Wang et al.~\cite{hybridxiushen} further introduced supervised contrastive learning into decoupled learning to improve the discriminative ability of image representations. Recently, Alshammari et al.~\cite{LTRweightbalancing} developed a sequential decoupled learning method with weight balancing strategies and achieved promising performance. 3) Ensembling: Handling different parts of the long-tailed data with multiple experts shows promising performance for long-tailed image recognition, which works essentially as ensembling by integrating these experts to obtain the optimal model over the entire dataset ~\cite{zhang2021test,xiang2020learning,cai2021ace}. Concretely, \cite{xiang2020learning} distilled multiple teacher models into a unified model, and each teacher focused on a relatively balanced group such as many-shot, medium-shot and few-shot classes. While for~\cite{wang2021longtailed}, it learned multiple distribution-aware experts by a dynamic expert routing module.

\subsection{Simplicity Bias in Neural Networks}

In the literature, simplicity bias (SB), \ie, the tendency of supervised neural networks to find simple patterns, was proposed to analyze the generalization of neural networks~\cite{simpBiasNIPs20,huh2021low}. In particular, in~\cite{simpBiasNIPs20}, it originally showed that neural networks trained with SGD are biased to learning the simplest predictive features in the data, while overlooking the complex but equally-predictive patterns. Inspired by this, Teney et al.~\cite{OOD} demonstrated that SB can be mitigated through a diversity constraint and explored how to improve out of distribution generalization by overcoming SB. In this paper, to our best knowledge, we are the first to investigate the SB problem in the long-tailed image recognition task, and propose methods that can effectively mitigate SB in long-tailed distribution learning.

\subsection{Self-Supervised Learning}

Self-supervised learning (SSL) has attracted increasing attention thanks to its ability to learn meaningful data representations without human annotation. It is capable of adopting self-defined pseudo labels as supervision and employs the learned representations for various downstream tasks, \eg, object detection, segmentation in computer vision. In particular, contrastive learning has recently become a dominant research area in SSL, which aims at learning augmentation-invariant representations. The basic idea thereof is contrasting the agreement between positive pairs, \eg, images under different augmentations, against those from negative pairs. More specifically, in SimCLR~\cite{chen2020simple}, a successful end-to-end model consisting of two encoders was proposed, where one encoder generates representations for positive samples and the other learns representations for negative samples. However, SimCLR required a large batch size to achieve satisfactory performance. The idea of memory bank~\cite{wu2018unsupervised} provides a potential solution for alleviating this limitation by pre-storing encodings of negative samples. Later, in MoCo~\cite{He_2020_CVPR}, it followed a dictionary look-up perspective for conducting contrastive learning by designing a dynamic dictionary with a queue and a moving-averaged encoder. SwAV~\cite{caron2020unsupervised} was proposed by utilizing a clustering algorithm to group similar feature together for similarity calculation. Very recently, \cite{tengyunipsworkshop} investigated SSL under the dataset imbalance setting and studied the problem of robustness to imbalanced training of self-supervised representations. In this paper, we empirically validate that SSL can learn more comprehensive features from input images, which can mitigate simplicity bias, especially for the tail data in long-tailed distribution. Moreover, we further propose a novel SSL method tailored for imbalanced data to boost the quality of representations learned from long-tailed data.

\begin{figure*}[t!]
	\centering
	\subfloat[\small {Examples of \emph{MNIST-CIFAR-LT}}]  {\includegraphics[width=0.95\textwidth]{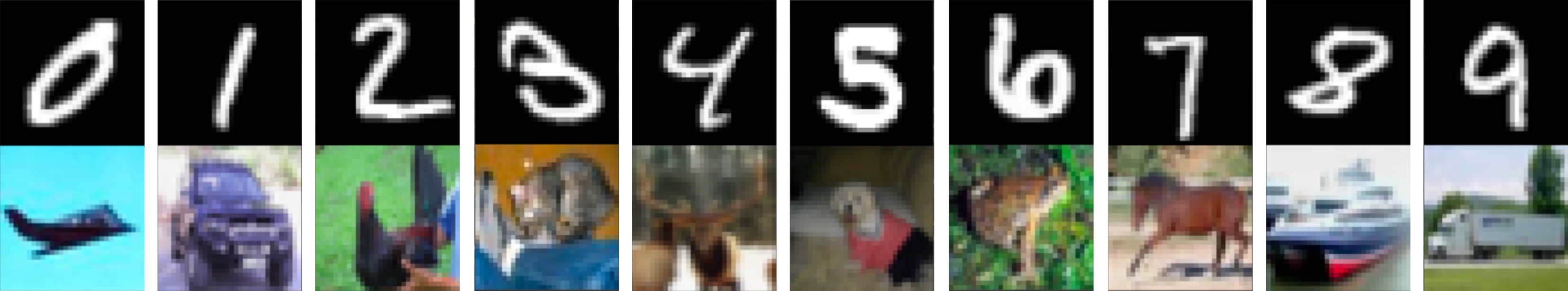} \label{fig:pre_exp_oriimg}}\\
	\subfloat[\small {Activations by cross-entropy}]  {\includegraphics[width=0.95\textwidth]{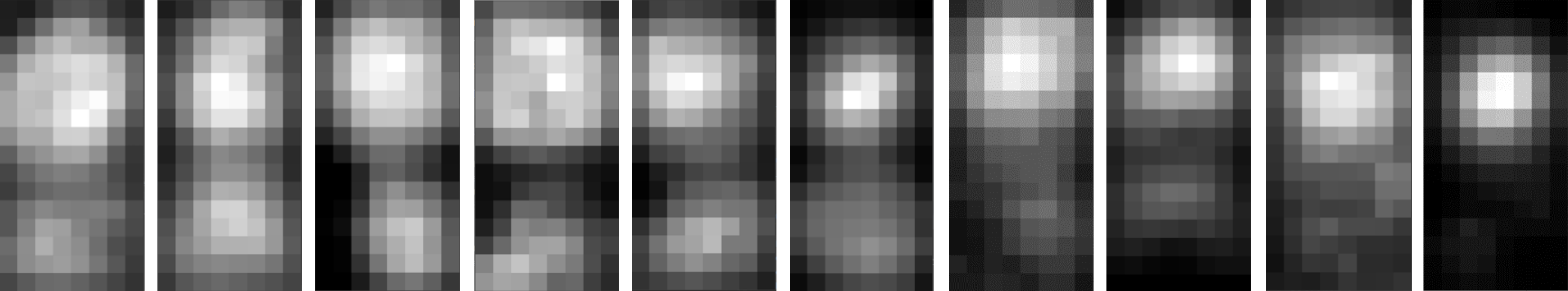} \label{fig:pre_exp_actiCE}}\\
	\subfloat[\small {Activations by our SSL baseline}]  {\includegraphics[width=0.95\textwidth]{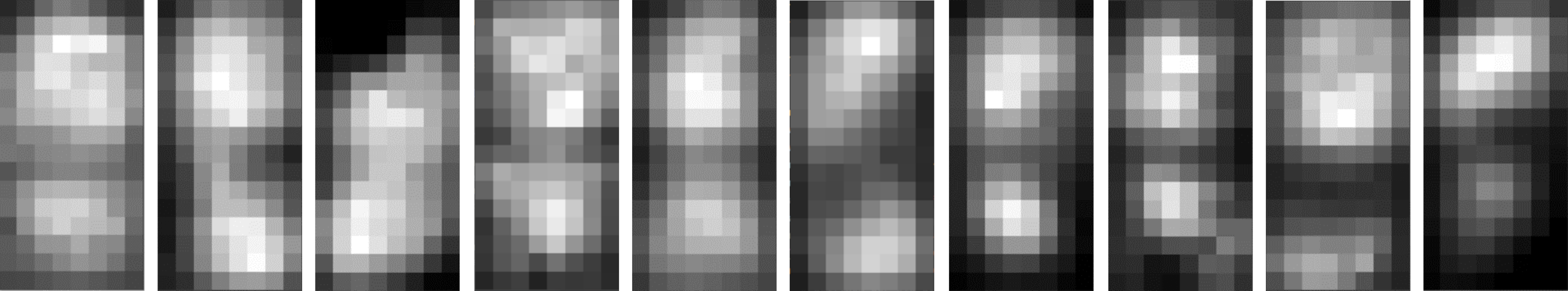} \label{fig:pre_exp_actiSSL}}
	\caption{\small Examples of our constructed \emph{MNIST-CIFAR-LT} dataset, and the corresponding activations obtained by traditional cross-entropy based supervised learning and our self-supervised baseline method. Note that, these ten samples are from head, medium, and tail data in the descending order of image quantity of the long-tailed distribution, respectively. The imbalance ratio is $100$.}
	\label{fig:pre_exp}
\end{figure*}

\section{The Pitfalls of Simplicity Bias in Long-Tailed Image Recognition}\label{sec:SB}

Simplicity bias (SB)~\cite{simpBiasNIPs20} is recently observed from neural networks trained with stochastic gradient descent, which shows the networks rely preferentially on few simple predictive features while ignoring more complex predictive features. The SB phenomenon partially explains the lack of robustness and generalization of deep neural networks in various vision problems, \eg, out of distribution~\cite{OOD}, confidence calibration~\cite{calibICML}, fooling examples~\cite{foolCVPR15}, etc. In this section, we investigate SB in the long-tailed image recognition task by conducting a series of preliminary experiments, as well as performing a simple baseline method for preliminary empirical verification.

\subsection{Preliminary Empirical Studies}

We present the dataset and preliminary results in the following, as well as discussing the investigation of simplicity bias in long-tailed image recognition.

\subsubsection{Dataset}

\begin{table*}[t!]
\centering
\small
\renewcommand\arraystretch{1.1}
\setlength{\tabcolsep}{4.5pt}
\caption{\small Preliminary results of classification accuracy (\%) on the constructed \emph{MNIST-CIFAR-LT} dataset with different imbalance ratios (\ie, 100, 10, and 1).}
\begin{tabular}{c|cccc|cccc|cccc}
\toprule
\multirow{2}{*}{\textsc{Test data}} & \multicolumn{4}{c|}{100}        & \multicolumn{4}{c|}{10}         & \multicolumn{4}{c}{1}          \\
\cline{2-13}
                           & \textsf{Head}  & \textsf{Medium} & \textsf{Tail}  & \textsf{All}   & \textsf{Head}  & \textsf{Medium} & \textsf{Tail}  & \textsf{All}   & \textsf{Head}  & \textsf{Medium} & \textsf{Tail}  & \textsf{All} \bigstrut[t]  \\
\hline
\emph{MNIST-CIFAR-LT}      & 99.90 & 99.40  & 98.50 & 99.33 & 99.90 & 99.80  & 99.63 & 99.79 & 99.87 & 99.93  & 99.90 & 99.90 \bigstrut[t]\\
\emph{MNIST-only}          & 99.60 & 98.40  & 94.90 & 97.83 & 99.55 & 98.93  & 98.23 & 98.97 & 99.72 & 99.47  & 98.96 & 99.42 \\
\emph{CIFAR-only}          & 61.67 & 59.00  & 32.16 & 52.02 & 66.10 & 69.80  & 68.16 & 67.83 & 64.40 & 80.50  & 84.46 & 75.25 \\
\hline
\emph{CIFAR-only (Oracle)} & 88.17 & 73.16 & 69.53 & 78.08 & 87.92 & 85.83 & 88.40 & 87.44 & 85.58 & 92.06 & 95.03 & 91.16 \\
\hline
\rowcolor[rgb]{ .851,  .851,  .851} Accuracy gap & -26.50 & -14.16 & -37.37 & -26.06 & -21.82 & -16.03 & -20.24 & -19.61 & -21.18 & -11.56 & -10.57 & -15.91 \\
\bottomrule                
\end{tabular}
\label{table:preMNCILT}
\end{table*}

In order to explicitly inspect the simplicity bias (SB) phenomenon on long-tailed image recognition, we construct a long-tailed toy dataset by vertically concatenating \emph{MNIST}~\cite{mnist} images onto \emph{CIFAR-10}~\cite{cifar} images. The dataset contains 10 classes, where each class is a combination of a unique \emph{MNIST} class and \emph{CIFAR-10} class\footnote{The constructed dataset has the following correspondences of the original classes of \emph{MNIST} and \emph{CIFAR-10}: ``\texttt{0-airplane}'', ``\texttt{1-automobile}'', ``\texttt{2-bird}'', ``\texttt{3-cat}'', ``\texttt{4-deer}'', ``\texttt{5-dog}'', ``\texttt{6-frog}'', ``\texttt{7-horse}'', ``\texttt{8-ship}'', and ``\texttt{9-truck}''.}, cf. Figure~\ref{fig:pre_exp_oriimg}. More importantly, these concatenated images follow a long-tailed distribution in terms of classes, where the number of images per class decreases monotonically from class with digit ``\texttt{0}'' to class with digit ``\texttt{9}''. The imbalance ratio\footnote{The imbalance ratio $\beta=\frac{N_{max}}{N_{min}}$, where $N_{max}$ and $N_{min}$ denote the number of training samples for the most and least frequent classes, respectively.} 
is varied by 100, 10 and 1, which are similar to the previous work~\cite{cb-focal,ldam}. We term this constructed dataset as \emph{MNIST-CIFAR-LT}. More specifically, its training and test splits comprise 50,000 and 10,000 images of size $3\times 64\times 32$, where the training set is long-tailed and the test set is class-balanced. Note that, the images in \emph{MNIST} are duplicated across three channels to match \emph{CIFAR} dimensions before concatenation.

\subsubsection{Preliminary results and discussions}

We first conduct standard supervised training, \ie, network training of ResNet-32~\cite{he2016deep} with cross-entropy, on \emph{MNIST-CIFAR-LT}. After the network converges, we visualize the activations of input \emph{MNIST-CIFAR-LT} images in the test set. Figure~\ref{fig:pre_exp_oriimg} shows ten example images from head, medium, and tail classes in the descending order of image quantity. From the visualization results in Figure~\ref{fig:pre_exp_actiCE} we can clearly observe that the MNIST parts are activated in all of the cases while the activations of CIFAR parts are reduced with decreasing number of images, or even nearly completely ignored. This observation indicates SB indeed happens because in this dataset digit patterns of \emph{MNIST} are simpler than the object patterns of \emph{CIFAR}. More importantly, the SB phenomenon becomes more significant on the tail data. For example, the ``\texttt{truck}'' is almost not activated in the ``\texttt{9-truck}'' example. This empirical study shows there exists correlation between SB and the amount of training data available to a specific class.

To inspect the SB in long-tailed image recognition quantitatively, we leave the training data of \emph{MNIST-CIFAR-LT} intact, while proposing different setups of test data, \ie, \emph{MNIST-CIFAR-LT}, \emph{MNIST-only}, and \emph{CIFAR-only}. \emph{MNIST-only}/\emph{CIFAR-only} means that we only use the MNIST/CIFAR part for testing. The results are reported in Table~\ref{table:preMNCILT}. As presented, compared with the testing accuracy on \emph{MNIST-CIFAR-LT}, the classification accuracy on \emph{MNIST-only} has a slight drop. However, the test data of \emph{CIFAR-only} shows a large overall accuracy reduction, which confirms the existence of SB again.

But, when further investigating the influence of SB on the tail data, we face a bias issue in the training set. As seen, when the imbalance ratio is 1, \ie, no long-tailed training distribution, the classification accuracy on \emph{CIFAR-only} of ``head'', ``medium'' and ``tail''\footnote{In fact, there are no ``head'', ``medium'' and ``tail'' classes in a balanced data distribution (\ie, the imbalance ratio is 1).} is 64.40\%, 80.50\%, and 84.46\%, respectively. There is a significantly large accuracy gap between these classes based on such a balanced training set. Speaking directly, on a balanced training set of the aforementioned \emph{MNIST-CIFAR-LT}, the CIFAR parts of ``tail'' (whose accuracy is 84.46\%) are easier for classification than the CIFAR parts of ``head'' (whose accuracy is 64.40\%). To gain a clearer and more quantitative understanding of how much tail classes suffer from SB in long-tailed distribution data, we create an \emph{oracle model} which trains on randomized blocks of MNIST parts and intact blocks of CIFAR parts in the constructed \emph{MNIST-CIFAR-LT} dataset by following~\cite{OOD}. Since the oracle trains on randomized blocks of MNIST, it is forced to only learn the patterns of CIFAR parts. Thus, the classification accuracy of such a model on the test set of \emph{CIFAR-only} will act as the oracle performance on head/medium/tail data, \ie, the results in the row of ``\emph{CIFAR-only (Oracle)}'' in Table~\ref{table:preMNCILT}.

In concretely, by comparing the results of ``\emph{CIFAR-only}'' and ``\emph{CIFAR-only (Oracle)}'', we can get the accuracy reduction gap caused by SB while removing the training bias as much as possible, cf. ``Accuracy gap'' in Table~\ref{table:preMNCILT}. Apparently, due to SB, the accuracy on the test set of \emph{CIFAR-only} under different imbalance ratio has dropped by at least 10\%, which again quantitatively confirms the influence of SB. More importantly, when comparing the accuracy gap results of different imbalance ratio horizontally, we find that as the long-tailed distribution imbalance increases, the tail data is more seriously affected by SB. For example, when the imbalance ratio is 1, the accuracy reduction gap caused by SB on ``head'' is $-$21.18\%, and the accuracy gap on ``tail'' is $-$10.57\%. With the increment of imbalance ratios, the accuracy reduction gap on head data grows to $-$21.82\% and $-$26.50\% with imbalance ratios of 10 and 100, respectively. It can be said that the accuracy reduction on head data is basically not affected by the long-tailed distribution ($-$21.18\% vs. $-$21.82\%/$-$26.50\%). However, the accuracy reduction gap on tail data increases dramatically from $-$10.57\% to $-$20.24\% and $-$37.37\% with imbalance ratios of 10 and 100, which is quite significant. On the other side, from the perspective of absolute accuracy comparison, when the imbalance ratio is 100, even if the CIFAR parts of ``tail'' are easier to classify due to the aforementioned training bias, its classification accuracy only achieves 32.16\% under the influence of more serious SB on tail data under the long-tailed distribution. While also, the classification accuracy of head data on the test set of \emph{CIFAR-only} is basically stable at about 60\% for different imbalance ratios.

From the aforementioned observations, we can have the conclusion that \emph{the tail data suffers more severely from SB}.

\subsection{A Baseline Method Incorporating Self-Supervised Learning}\label{sec:SSLbaseline}

As aforementioned, SB does appear in long-tailed image recognition and has a more dramatic effect on the tail data. To alleviate this problem, we propose to use self-supervised learning (SSL) to learn more comprehensive features/patterns, especially complex features/patterns.

As preliminary verification, we design an experiment with a baseline method incorporating SSL. Concretely, we utilize different approaches to pre-train a ResNet-32~\cite{he2016deep} network on the \emph{long-tailed CIFAR-100}~\cite{ldam} dataset with imbalance ratio of 100, and then fix the backbone but fine-tune the classifier via class-balanced sampling. In this way, we compare the generalization ability of the learned image representations, which can implicitly reflect the degree of SB~\cite{simpBiasNIPs20}.

\begin{table}[t]
\centering
\small
\caption{\small Preliminary results of classification accuracy (\%) on \emph{long-tailed CIFAR-100}~\cite{ldam} with imbalance ratio of 100. ``CE'' represents cross-entropy, and ``SSL'' denotes self-supervised learning.}
\begin{tabular}{c|cccc}
\toprule
\textsc{Approach}    & \textsf{Head}           & \textsf{Medium}         & \textsf{Tail}           & \textsf{All}            \\
\hline
CE        & 62.83          & \textbf{56.85} & 45.36          & 55.50     \bigstrut[t]     \\
Ours (CE w. SSL) & \textbf{64.29} & 56.37          & \textbf{48.20} & \textbf{56.69} \\
SSL       & 31.23          & 30.43          & 24.37          & 28.89         \\
\bottomrule                
\end{tabular}
\label{table:preSSL}
\end{table}

\begin{figure}[t!]
	\centering
	\subfloat[\small Learning curves with cross-entropy]  {\includegraphics[width=0.9\columnwidth]{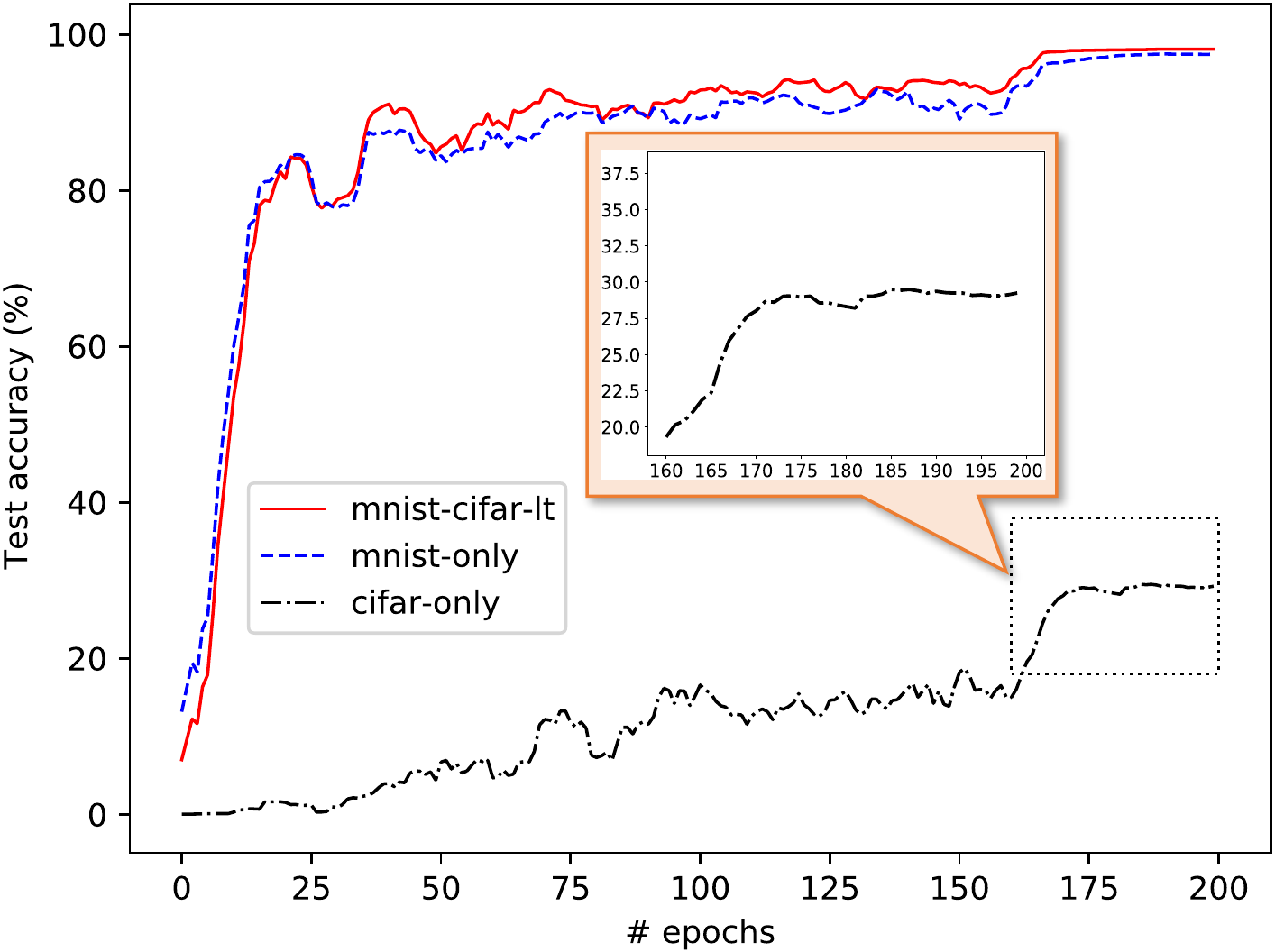} \label{fig:curve1}}\\
	\subfloat[\small Learning curves with our SSL baseline]  {\includegraphics[width=0.9\columnwidth]{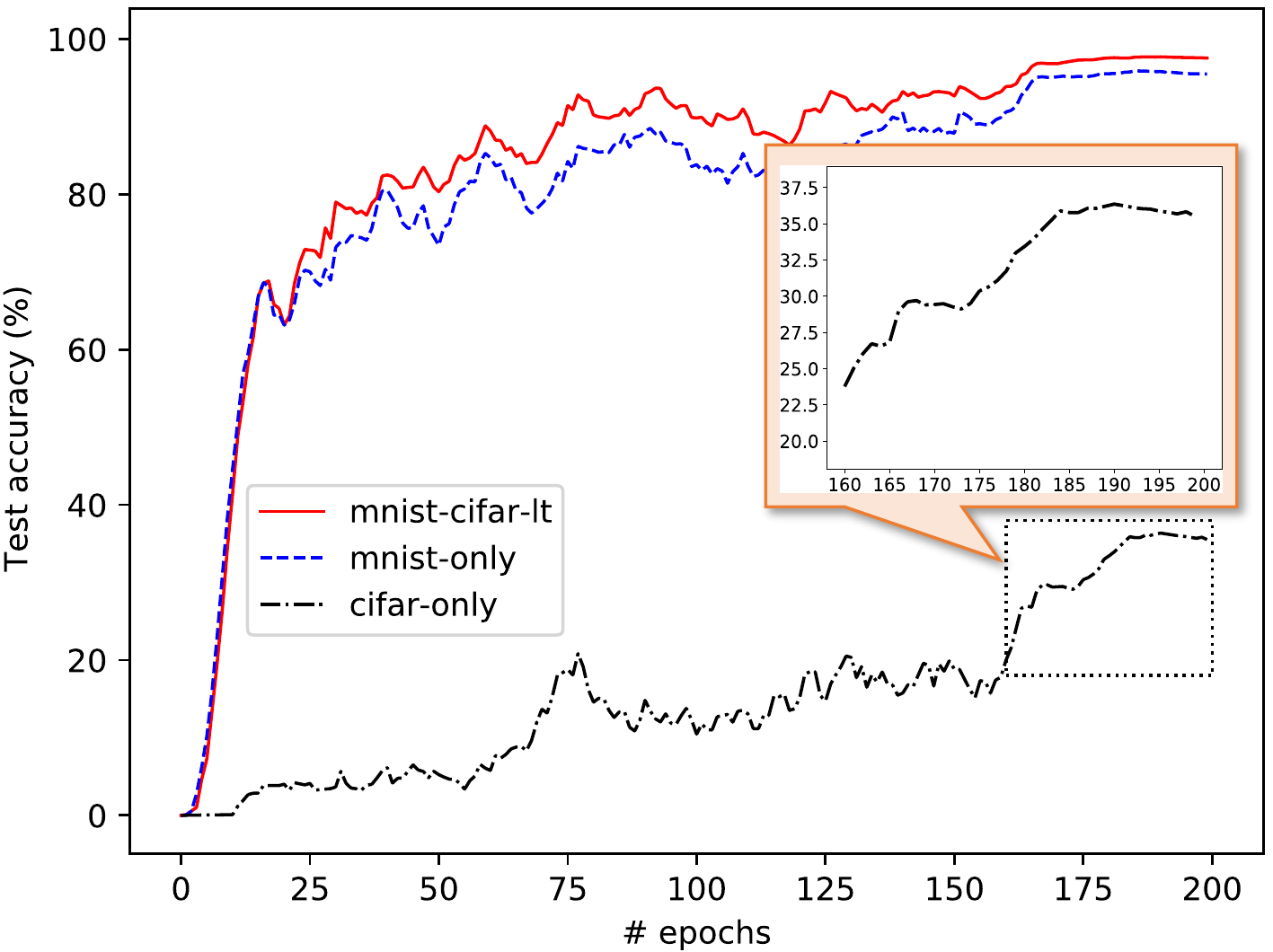} \label{fig:curve2}}
	\caption{\small Learning curves of test accuracy (\%) vs training epochs (from 0 to 200) on the test set of \emph{MNIST-CIFAR-LT}, \emph{MNIST-only} and \emph{CIFAR-only}.}
	\label{fig:curve}
\end{figure}

These pre-trained approaches as well as the classification accuracy obtained thereof are presented in Table~\ref{table:preSSL}. More specifically, when comparing to pre-training with cross-entropy, we further equip self-supervised learning into pre-training by applying MoCo-v2~\cite{mocov2} in a multi-task learning framework, which is denoted by ``Ours (CE w. SSL)'', as our SSL baseline in Table~\ref{table:preSSL}. As shown, combining SSL, the classification accuracy obtains an obvious improvement on the tail data (about 3\% improvements). In addition, we also report the results of only using SSL for pre-training, and we can see that it achieves inferior classification accuracy. These phenomena indicate that supervised learning and self-supervised learning are complementary to each other in this case.

Furthermore, we also perform the SSL baseline method on \emph{MNIST-CIFAR-LT} and show the learning curves with cross-entropy and our SSL baseline in Figure~\ref{fig:curve}. As shown, after applying SSL, the final classification accuracy of \emph{CIFAR-only} has been improved. These observations in Figure~\ref{fig:curve} show that SB has been greatly alleviated by the self-supervised fashion. Also, we visualize the activations by performing our SSL baseline in Figure~\ref{fig:pre_exp_actiSSL}. It can also be clearly observed that the network tends to attend more complex patterns of the CIFAR part thanks to SSL.


\section{Methodology}\label{sec:approach}

Inspired by the preliminary experiments, we propose a simple but effective method, \ie, Triple-Level Self-Supervised Learning ({3LSSL}), to incorporate SSL into long-tailed image recognition. It leverages SSL by triple different levels to enhance the learning of predictive complex patterns, providing the potential to overcome the severe SB on tail data.

\begin{figure*}[t!]
	\centering
	{\includegraphics[width=0.55\textwidth]{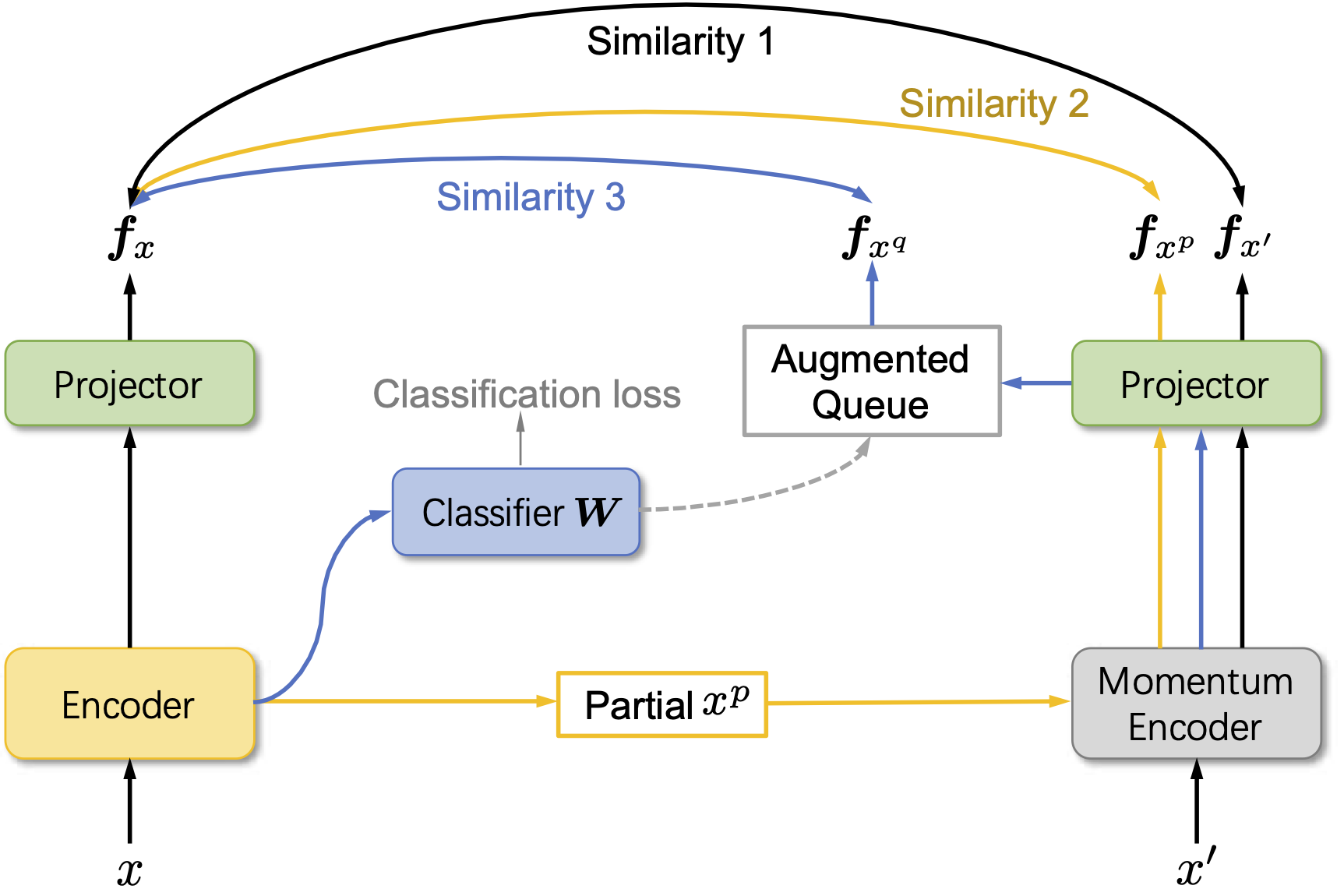}}
	\vspace{-0.75em}
	\caption{\small Overall framework of the proposed Triple-Level Self-Supervised Learning (3LSSL) method. Note that, ``Similarity 1'' corresponds to the holistic-level SSL, which calculates the cosine similarity between the embeddings $\bm{f}_x$ and $\bm{f}_{x'}$ w.r.t. the input $x$ and its holistic counterpart $x'$, respectively. ``Similarity 2'' represents the partial-level SSL, which calculates the similarity between $\bm{f}_x$ and $\bm{f}_{x^p}$ (w.r.t. a partial view of $x$). ``Similarity 3'' is the augmented-level SSL, which calculates the similarity of $\bm{f}_x$ with $\bm{f}_{x^q}$, where $\bm{f}_{x^q}$ is the augmented embeddings w.r.t $x$ based on an augmented queue derived from a re-balanced classifier.}
	\label{fig:method}
\end{figure*}

\subsection{Holistic-Level SSL}

We follow the classic SSL framework, \eg, MoCo~\cite{He_2020_CVPR,mocov2}, to realize the holistic-level self-supervised learning in our 3LSSL method. More specifically, for an image $\mathcal{I}$, we augment it in different ways and obtain $x$ and $x'$ as inputs of the encoder and momentum encoder in SSL. Then, after performing two projectors, the image representations of $x$ and $x'$ are obtained, \ie, $\bm{f}_{x}$ and $\bm{f}_{x'}$. Different from previous SSL methods using InfoNCE as the loss function~\cite{He_2020_CVPR,mocov2}, we employ $\ell_2$-normalization upon $\bm{f}_{x}$ and $\bm{f}_{x'}$ and calculate the cosine similarity to drive the network training, which is as follows:
\begin{equation}
\mathcal{L}_{sim_{1}} = \bm{f}_{x}^\top \cdot \bm{f}_{x'}\,,
\end{equation}
where we omit $\ell_2$-normalization for simplification. The holistic-level SSL is presented as the flow with black arrows in Figure~\ref{fig:method}, which aims to enhance the learning of comprehensive patterns in the holistic level w.r.t. $x$.

Regarding the update of momentum encoder in our method, we follow~\cite{He_2020_CVPR} by conducting
\begin{equation}
\label{eq:MomLTSB-Cls}
\theta_k^\top=\alpha\theta_k^{t-1}+(1-\alpha)\theta_q^{t-1} \,,
\end{equation}
where $t$ represents the current step, and $t-1$ means the previous step. $\theta_q$ and $\theta_k$ represent the parameters of encoder and momentum encoder, respectively. $\alpha$ is the updating rate of the momentum encoder, which is fixed as $0.999$.

\subsection{Partial-Level SSL}

Through the visualized activations in the preliminary experiments on \emph{MNIST-CIFAR-LT}, cf. Figure~\ref{fig:pre_exp_actiSSL} vs. Figure~\ref{fig:pre_exp_actiCE}, it verifies that holistic SSL (\ie, the SSL baseline in Section~\ref{sec:SSLbaseline}) is able to make models learn more sufficient patterns (\ie, alleviating simplicity bias) than vanilla models to some extent. However, there are still some important but complex image regions that are ignored due to SB, \eg, the upper part of the ship or the head of the truck in Figure~\ref{fig:pre_exp_actiSSL}. To solve the problem that holistic-level SSL might ignore complex patterns, we develop a partial-level SSL by leveraging a masking manner to force the models to learn more comprehensive information from the partial level w.r.t. the whole image. The partial-level SSL is illustrated as the flow with yellow arrows in Figure~\ref{fig:method}.

Concretely, for the input $x$, we can obtain its deep feature map $\bm{X}\in \mathbb{R}^{h\times w\times d}$ by performing the encoder, where $h$, $w$ and $d$ represent the height, width and depth of the feature map, respectively. Then, we employ an off-the-shelf localization approach, \ie, Class Activation Mapping (CAM)~\cite{CAM}, upon $\bm{X}$ with the re-balanced classifier $\bm{W}$ (elaborated lated in Section~\ref{sec:augmentSSL}) to indicate the class-specific discriminative image regions used by CNNs to identify that class. It could also expose the implicit attention of CNNs w.r.t. a target class on an image. Obeying the notations in~\cite{CAM}, we conduct CAM by projecting back the weights of the output layer on to the convolutional feature maps as
\begin{equation}
{M}_c(i,j) = \sum_k w_k^c f_k(i,j) \,,
\end{equation}
where $w_k^c$ is the weight of $\bm{W}$ w.r.t. class $c$ for neural unit $k$ in the last convolutional layer of the encoder, and $f_k(i,j)$ represent the activation of unit $k$ at spatial location $(i,j)$. Thus, $M_c(i,j)$ directly indicates the importance of the activation at  spatial grid $(i,j)$ leading to the classification of an image to class $c$, and $\bm{M}_c\in \mathbb{R}^{h\times w}$ corresponds to the so-called class activation map. Intuitively, $\bm{M}_c$ is simply a weighted linear sum of the presence of these visual patterns at different spatial locations. By further simply upsampling $\bm{M}_c$ to the size of the input image, it can identify the image regions most relevant to the particular class. Therefore, to achieve the goal of our partial-level SSL as aforementioned, we deploy $\bm{M}_c$ as a mask for filtering out those class-specific image regions that are most likely to be attended by deep networks, and force it to learn more comprehensive information from the remaining image regions. In our implementation, we normalize $\bm{M}_c$ by
\begin{equation}
\hat{{M}_c}(i,j) = \frac{M_c(i,j)-\min(\bm{M}_c)}{\max(\bm{M}_c)-\min(\bm{M}_c)}\,,
\end{equation}
where $\hat{{M}_c}(i,j)$ forms the normalized mask $\hat{\bm{M}_c}$ whose values are in the range of $\left[0,1\right]$. $\min(\bm{M}_c)$ and $\max(\bm{M}_c)$ are the minimum and maximum value of $\bm{M}_c$, respectively.

After that, the input w.r.t. the partial-level SSL $x^p$ is obtained by
\begin{equation}
x^p = x \odot \left(\bm{J}-{\rm upsample}(\hat{\bm{M}_c}) \right) \,,
\end{equation}
where $\bm{J}$ is the unit matrix with all $1$ elements, $\odot$ is the element-wise product, and ${\rm upsample}(\cdot)$ represents the upsampling operation to the size of original image resolution. Thus, as shown by the yellow arrows in Figure~\ref{fig:method}, the embedding of $x^p$, \ie, $\bm{f}_{x^p}$, is obtained through the momentum encoder followed by the projector. Similar to the loss in the holistic-level SSL, we calculate the similarity between the $\ell_2$-normalized $\bm{f}_{x^p}$ and $\bm{f}_{x}$ by
\begin{equation}
\mathcal{L}_{sim_{2}} = \bm{f}_{x}^\top \cdot \bm{f}_{x^p}\,,
\end{equation}
to realize the partial-level SSL in our method.

\subsection{Augmented-Level SSL}\label{sec:augmentSSL}

To further consider the severe SB on tail data, we enhance the similarity-based representation learning of tail classes by borrowing pseudo samples from other classes informed by classifier outputs, which is termed as the augmented-level SSL. As the flow of blue arrows shown in Figure~\ref{fig:method}, we first build a re-balanced classifier $\bm{W}$ upon $x$ and then produce an augmented queue containing augmented samples $x^q$ from $x'$ for achieving the aforementioned goal.

More specifically, we equip a supervised component by training a classifier upon the encoder's output for recognizing $x$. It also complies with the conclusion that supervised learning and self-supervised learning complement each other in the case of long-tailed recognition, cf. Table~\ref{table:preSSL}. The classification loss w.r.t. $\bm{W}$ hereby is the cross-entropy loss function on the ground truth and the calibrated predicted logits which are obtained by adding up re-balanced factors of $\{\ln (N_c/\sum_j N_j)\}$ to original predicted logits (where $N_c$ is the number of samples in class $c$).

To enhance the similarity-based representation learning of the tail data $x$, in the augmented-level SSL, we use $\bm{W}$ to predict the class affiliation $c'$ of $x'$ and select samples from class $c'$ as pseudo positives of $x$. If we denote the embedding of such pseudo positive as $\bm{f}_{c'}$, we add another similarity-based loss on such a pair $(\bm{f}_{x}, \bm{f}_{c'})$. By doing this, we augment the tail classes by providing chances to see more semantic-relevant samples for representation learning and consequently learn features that generalize better. 

Furthermore, in order to achieve better accuracy and make the training more stable, we develop an augmented queue strategy to pre-store embeddings $\bm{f}_{y}$ (\ie, a general form of $\bm{f}_{c'}$ as aforementioned) and their predicted pseudo labels $y$. In concretely, for such a size-$L$ augmented queue $\left\{(\bm{f}_{y^l}^l,y^l)\right\}_{l=1}^L$, we initially set the feature vectors as $\bm{0}$ and their predictions as $-1$. After obtaining a pseudo-label $y$ of $x'$ at each time step, the embedding-pseudo label pair in the earliest time step will be removed, and it adds the newly combing embedding-pseudo label pair to the queue. Then, we collect all embeddings associated with class $c'$ in the augmented queue to obtain an aggregated feature embedding, which is formulated by
\begin{equation}
\bm{f}_{x^q} = \frac{1}{|\Omega|}\sum_{l\in\Omega} \bm{f}_{y^l}^l\,,
\end{equation}
where $\Omega=\{l|y^l==c'\}$. After that, the augmented-level SSL is optimized by computing the similarity between the $\ell_2$-normalized $\bm{f}_{x^q}$ and $\bm{f}_{x}$, which also follows the holistic- and partial-level SSL:
\begin{equation}
\mathcal{L}_{sim_{3}} = \bm{f}_{x}^\top \cdot \bm{f}_{x^q}\,.
\end{equation}

Overall, the final loss function of our 3LSSL method is optimized by
\begin{equation}
\mathcal{L}=\mathcal{L}_{sim_{1}}+\mathcal{L}_{sim_{2}}+\mathcal{L}_{sim_{3}}+\mathcal{L}_{cls}\,,
\end{equation}
where $\mathcal{L}_{cls}$ denotes the classification loss w.r.t. $\bm{W}$. All trade-off parameters are $1$ for all experiments, which reveals the good potential of our method in practice and its simplicity.

\section{Experiments}\label{sec:experiments}

In this section, we present the empirical settings, implementation details, main results, ablation studies and also some qualitative analyses of visualization.

\subsection{Datasets, Settings, and Implementation Details}

We conduct experiments on five long-tailed image recognition benchmark datasets, including \emph{long-tailed CIFAR-10}/\emph{long-tailed CIFAR-100}~\cite{cifar}, \emph{ImageNet-LT}~\cite{oltr}, \emph{Places-LT}~\cite{zhou2017places} and \emph{iNaturalist 2018}~\cite{van2018inaturalist}. Furthermore, we also report the results on the constructed \emph{MNIST-CIFAR-LT} dataset. For quantitative comparisons, we evaluate the methods on the corresponding balanced validation datasets, and report the top-1 accuracy over all classes, which is denoted as ``\textsf{All}''. Besides, we follow~\cite{decouple,oltr} to split the classes into three subsets and report the average accuracy in these three subsets, \ie, ``\textsf{Many}''-shot ($>$100 images), ``\textsf{Medium}''-shot (20$\sim$100 images), and ``\textsf{Few}''-shot ($<$20 images), which are also termed as head, medium and tail data, respectively.

\begin{itemize}
\item{\textbf{\emph{Long-tailed CIFAR-10/long-tailed CIFAR-100}}}: Both \emph{CIFAR-10} and \emph{CIFAR-100}~\cite{cifar} contain 60,000 images, 50,000 for training and 10,000 for validation with category number of 10 and 100, respectively. For fair comparisons, we use the long-tailed versions of CIFAR datasets as the same as those used in~\cite{ldam} with controllable degrees of data imbalance. We use an imbalance factor $\beta$ to describe the severity of the long tail problem with the number of training samples for the most frequent class and the least frequent class, \eg, $\beta=\frac{N_{max}}{N_{min}}$. Imbalance factors we use in experiments are $10$, $50$ and $100$.


\item{\textbf{\emph{ImageNet-LT}}}: The ImageNet-LT dataset~\cite{oltr} is a long-tailed version of the original ImageNet-2012~\cite{ILSVRC15}, which is constructed by sampling a subset following the Pareto distribution with the power value as $6$. It contains 115.8K large-scale images from 1,000 categories. The number of training samples for each class ranges from 5 to 1,280. The validation set is the same as ImageNet 2012.


\item{\textbf{\emph{Places-LT}}}: The Places-LT dataset is also a large-scale long-tailed dataset artificially created from the balanced Places-2~\cite{zhou2017places} dataset. It contains 184.5K images from 365 diverse scene categories. The distribution of labels in the training set is also extremely long-tailed, where the sample number of each class ranges from 5 to 4,980.

\item{\textbf{\emph{iNaturalist 2018}}}: iNaturalist 2018~\cite{van2018inaturalist} is a large-scale real-world dataset with 437.5K images from 8,142 categories. It naturally follows a severe long-tailed distribution with an imbalance factor of 512. Besides the extreme imbalance, it also faces the fine-grained problem~\cite{wei2019deep,pcm,scda}. In this paper, the official splits of training and validation images are utilized for fair comparisons.
\end{itemize}

\begin{table*}[t]
\centering
\small
\renewcommand\arraystretch{1.1}
\setlength{\tabcolsep}{13.5pt}
\caption{\small Comparisons of classification accuracy (\%) on \emph{Long-tailed CIFAR-10} and \emph{Long-tailed CIFAR-100} with ResNet-32. The highest accuracy is marked in bold.}
\begin{tabular}{c|ccc|ccc}
\toprule
\multirow{2}{*}{\textsc{Method}}           & \multicolumn{3}{c|}{\emph{Long-tailed CIFAR-10}} & \multicolumn{3}{c}{\emph{Long-tailed CIFAR-100}} \\
                                           & 100            & 50             & 10            & 100            & 50             & 10             \\
\hline
LDAM-DRW~\cite{ldam}                       & 77.0           & 81.0           & 88.2          & 42.0           & 46.6           & 58.7 \bigstrut[t] \\
BBN~\cite{bbn}                             & 79.8           & 82.2           & 88.3          & 42.6           & 47.0           & 59.1           \\
UniMix~\cite{xu2021towards}                & 82.8           & 84.3           & 89.7          & 45.5           & 51.1           & 61.3           \\
RIDE (3 experts)~\cite{wang2021longtailed} & --             & --             & --            & 48.0           & 51.7           & 61.8           \\
DiVE~\cite{He_2021_ICCV}                   & --             & --             & --            & 45.4           & 51.1           & 62.0           \\
SSD~\cite{li2021self}                      & --             & --             & --            & 46.0           & 50.5           & 62.3           \\
MiSLAS~\cite{calibrationcvpr2021}          & 82.1           & 85.7           & 90.0          & 47.0           & 52.3           & 63.2           \\
GCL~\cite{GCLcvpr22}                       & 82.7           & 85.5           & --            & 48.7           & 53.6           & --             \\
DRO-LT~\cite{Samuel_2021_ICCV}             & --             & --             & --            & 47.3           & 57.6           & 63.4           \\
SADE~\cite{SADEnips22}                     & --             & --             & --            & 49.8           & 53.9           & 63.6           \\
ResLT~\cite{ResLTPAMI}                     & 82.4           & 85.2           & 89.7          & 49.7           & 54.5           & 63.7           \\
PaCo~\cite{Cui_2021_ICCV}                  & --             & --             & --            & 52.0           & 56.0           & 64.2           \\
BCL~\cite{BCLcvpr22}                       & 84.3           & 87.2           & 91.1          & 51.9           & 56.6           & 64.9           \\
\hline
\textbf{Our 3LSSL}                         & \textbf{85.2}  & \textbf{88.2}  & \textbf{92.1} & \textbf{54.6}  & \textbf{58.5}  & \textbf{66.2} \\
\bottomrule           
\end{tabular}
\label{table:resultsCIFARLT}
\end{table*}

\begin{table*}[t]
\centering
\small
\renewcommand\arraystretch{1.1}
\caption{\small Comparisons of classification accuracy (\%) on \emph{ImageNet-LT} with ResNet-50 and ResNeXt-50. The highest accuracy is marked in bold.}
\begin{tabular}{c|cccc|cccc}
\toprule
\multirow{2}{*}{\textsc{Method}}           & \multicolumn{4}{c|}{ResNet-50}                                 & \multicolumn{4}{c}{ResNeXt-50}                                \\
                                           & \textsf{Many} & \textsf{Med.} & \textsf{Few}  & \textsf{All}  & \textsf{Many} & \textsf{Med.} & \textsf{Few}  & \textsf{All}  \\
\hline
MiSLAS~\cite{calibrationcvpr2021}          & 61.7          & 51.3          & 35.8          & 52.7          & --            & --            & --            & --    \bigstrut[t] \\
ResLT~\cite{ResLTPAMI}                     & --            & --            & --            & --            & 63.0          & 50.5          & 35.5          & 53.0          \\
DiVE~\cite{He_2021_ICCV}                   & --            & --            & --            & --            & 64.1          & 50.4          & 31.5          & 53.1          \\
DRO-LT~\cite{Samuel_2021_ICCV}             & 64.0          & 49.8          & 33.1          & 53.5          & --            & --            & --            & --            \\
SSD~\cite{li2021self}                      & --            & --            & --            & --            & 66.8          & 53.1          & 35.4          & 56.0          \\
GCL~\cite{GCLcvpr22}                       & --            & --            & --            & 54.9          & --            & --            & --            & --            \\
RIDE (4 experts)~\cite{wang2021longtailed} & 66.2          & 52.3          & 36.5          & 55.4          & 68.2          & 53.8          & 36.0          & 56.8          \\
BCL~\cite{BCLcvpr22}                       & --            & --            & --            & 56.0          & 67.9          & 54.2          & 36.6          & 57.1          \\
RIDE (3 experts)+CMO~\cite{CMOcvpr22}      & 66.4          & 53.9          & 35.6          & 56.2          & --            & --            & --            & --            \\
PaCo~\cite{Cui_2021_ICCV}                  & 65.0          & 55.7          & 38.2          & 57.0          & 67.5          & 56.9          & 36.7          & 58.2          \\
SADE~\cite{SADEnips22}                     & --            & --            & --            & --            & 66.5          & 57.0          & \textbf{43.5}          & 58.8          \\
\hline
\textbf{Our 3LSSL}                         & \textbf{68.5} & \textbf{57.6} & \textbf{38.3} & \textbf{59.1} & \textbf{70.0} & \textbf{57.8} & {38.7} & \textbf{59.9} \\
\bottomrule           
\end{tabular}
\label{table:resultsImageNetLT}
\end{table*}

\begin{table}[t]
\centering
\small
\renewcommand\arraystretch{1.1}
\caption{\small Comparisons of classification accuracy (\%) on \emph{Places-LT} with ResNet-152. The highest accuracy is marked in bold.}
\begin{tabular}{c|cccc}
\toprule
\textsc{Method}                   & \textsf{Many} & \textsf{Med.} & \textsf{Few} & \textsf{All} \\
\hline
GistNet~\cite{gistnet21}          & 42.5          & 40.8          & 32.1         & 39.6   \bigstrut[t] \\
MiSLAS~\cite{calibrationcvpr2021} & 39.6          & 43.3          & \textbf{36.1}         & 40.4         \\
GCL~\cite{GCLcvpr22}              & --            & --            & --           & 40.6         \\
SADE~\cite{SADEnips22}            & --            & --            & --           & 40.9         \\
ResLT~\cite{ResLTPAMI}            & 40.3          & 44.4          & 34.7         & 41.0         \\
PaCo~\cite{Cui_2021_ICCV}         & 37.5          & \textbf{47.2}          & 33.9         & 41.2         \\
\hline
\textbf{Our 3LSSL}                &         \textbf{42.6}      &        44.5       &      34.3        &   \textbf{42.0}  \\
\bottomrule           
\end{tabular}
\label{table:resultsPlacesLT}
\end{table}

\begin{table}[t]
\centering
\small
\renewcommand\arraystretch{1.1}
\caption{\small Comparisons of classification accuracy (\%) on \emph{iNaturalist 2018} with ResNet-50. The highest accuracy is marked in bold.}
\begin{tabular}{c|cccc}
\toprule
\textsc{Method}                            & \textsf{Many} & \textsf{Med.} & \textsf{Few} & \textsf{All} \\
\hline
DiVE~\cite{He_2021_ICCV}                   & 70.6          & 70.0          & 67.6         & 69.1  \bigstrut[t] \\
BBN~\cite{bbn}                             & --            & --            & --           & 69.6         \\
DRO-LT~\cite{Samuel_2021_ICCV}             & --            & --            & --           & 69.7         \\
ResLT~\cite{ResLTPAMI}                     & 68.5          & 69.9          & 70.4         & 70.2         \\
SSD~\cite{li2021self}                      & --            & --            & --           & 71.5         \\
MiSLAS~\cite{calibrationcvpr2021}          & 70.4          & 72.4          & 73.2         & 71.6         \\
BCL~\cite{BCLcvpr22}                       & --            & --            & --           & 71.8         \\
GCL~\cite{GCLcvpr22}                       & --            & --            & --           & 72.0         \\
RIDE (4 experts)~\cite{wang2021longtailed} & 70.9          & 72.4          & 73.1         & 72.6         \\
RIDE (3 experts)+CMO~\cite{CMOcvpr22}      & 68.7          & 72.6          & 73.1         & 72.8         \\
SADE~\cite{SADEnips22}                     & --            & --            & --           & 72.9         \\
PaCo~\cite{Cui_2021_ICCV}                  & 70.3          & 73.2          & 73.6         & 73.2         \\
\hline
\textbf{Our 3LSSL}                         &      \textbf{71.0}         &      \textbf{75.7}         &   \textbf{76.2}          &   \textbf{75.8}     \\
\bottomrule           
\end{tabular}
\label{table:resultsiNat}
\end{table}

In our 3LSSL, the encoder and the momentum encoder are of the same architecture with the same parameter initialization. We follow previous work, \eg, \cite{bbn,xu2021towards,calibrationcvpr2021,GCLcvpr22,SADEnips22,ResLTPAMI,Cui_2021_ICCV,BCLcvpr22}, to employ the corresponding backbone deep model to realize the encoders for fair comparisons. The two projectors are also of the same structure which consists of a two-layer perceptron head (hidden layer 2048-d, with ReLU). Both projectors have the same parameter initialization. While, during network training, two encoders and two projectors update its own parameters independently. The classifier component in our 3LSSL is a linear classification layer. More specifically, for fair comparisons to prior art, we follow the protocol in~\cite{bbn,xu2021towards,calibrationcvpr2021,GCLcvpr22,SADEnips22,ResLTPAMI,Cui_2021_ICCV,BCLcvpr22} by employing ResNet-32~\cite{he2016deep} as the backbone for \emph{long-tailed CIFAR-10/long-tailed CIFAR-100}, ResNet-50~\cite{he2016deep} and ResNeXt-50~\cite{xie2017aggregated} for \emph{ImageNet-LT}, ResNet-152~\cite{he2016deep} for \emph{Places-LT}, and ResNet-50~\cite{he2016deep} for \emph{iNaturalist 2018}. Regarding other empirical settings, we follow the recent state-of-the-art, \eg,~\cite{BCLcvpr22,Cui_2021_ICCV}, to conduct model training. Regarding the size $L$ of the augmented queue, we set $L=1024$ for \emph{long-tailed CIFAR} and $L=8192$ for large-scale \emph{ImageNet-LT}, \emph{Places-LT} and \emph{iNaturalist 2018}. All experiments are conducted with four GeForce RTX 3090 Ti GPUs.

\subsection{Main Results}

\subsubsection{Comparisons on Benchmark Long-Tailed Recognition Datasets}

We report the classification accuracy on \emph{long-tailed CIFAR-10}/\emph{long-tailed CIFAR-100}, \emph{ImageNet-LT}, \emph{Places-LT} and \emph{iNaturalist} in Table~\ref{table:resultsCIFARLT}, Table~\ref{table:resultsImageNetLT}, Table~\ref{table:resultsPlacesLT}, and Table~\ref{table:resultsiNat}, respectively. On these benchmarks, our 3LSSL method is able to consistently obtain significant accuracy boost over state-of-the-arts.

\textbf{Comparisons on \emph{long-tailed CIFAR}}. In Table~\ref{table:resultsCIFARLT}, we compare our method with state-of-the-art methods with different imbalance ratios (\ie, $100$, $50$, $10$) on the \emph{long-tailed CIFAR} datasets. It is obviously to see that the proposed 3LSSL method surpasses the competing methods, including the effective ResLT~\cite{ResLTPAMI}, PaCo~\cite{Cui_2021_ICCV}, BCL~\cite{BCLcvpr22}, etc. In particular, our method achieves 2.7\%, 1.9\%, 1.3\% improvement over BCL~\cite{BCLcvpr22} on \emph{long-tailed CIFAR-100} with the imbalance ratio of $100$, $50$, $10$, respectively.

\textbf{Comparisons on \emph{ImageNet-LT}}. \emph{ImageNet-LT} is the most basic large-scale dataset in the comparisons of long-tailed recognition experiments. Different backbones are also performed on this dataset for generalization ability testing. We present the classification accuracy results of ResNet-50 and ResNeXt-50 on \emph{ImageNet-LT} in Table~\ref{table:resultsImageNetLT}. As shown, equipped with ResNet-50, our 3LSSL obtains 59.1\% (having 2.1\% improvement over PaCo~\cite{Cui_2021_ICCV}). Equipped with ResNeXt-50, our method achieves 59.9\%, which still outperforms the previous best result 58.8\% from SADE~\cite{SADEnips22} by 1.1\%.

\textbf{Comparisons on \emph{Places-LT}}. The comparisons of classification results on \emph{Places-LT} are summarized in Table~\ref{table:resultsPlacesLT}. We can observe that the improvement of the previous methods on this dataset is relatively small, \eg, 0.2$\sim$0.3\%. This also illustrates the challenges of \emph{Places-LT}. In contrast, our method achieves a relatively large accuracy boost of 0.8\% over PaCo~\cite{Cui_2021_ICCV}, which proves the effectiveness of the proposed 3LSSL method.

\textbf{Comparisons on \emph{iNaturalist 2018}}. The \emph{iNaturalist 2018} dataset is a natural large-scale long-tailed dataset. Therefore, the comparisons on this dataset, especially the comparisons on the tail data, can better illustrate the effectiveness of the methods. As shown in Table~\ref{table:resultsiNat}, our 3LSSL again achieves the best classification accuracy, \ie, 75.8\%, which significantly outperforms SADE~\cite{SADEnips22} and PaCo~\cite{Cui_2021_ICCV} by 2.9\% and 2.6\%, respectively. More importantly, on tail data (cf. the results of ``\textsf{Few}'' in Table~\ref{table:resultsiNat}), our method still obtains 2.6\% improvement over PaCo~\cite{Cui_2021_ICCV}.

\subsubsection{Comparisons on the Constructed \emph{MNIST-CIFAR-LT} Dataset}

We report the classification accuracy of our method on the constructed \emph{MNIST-CIFAR-LT} dataset with different imbalance ratios in Table~\ref{table:MNCILT}. The comparison of the results on this dataset requires to be combined with Table~\ref{table:preMNCILT} at the same time. By comparisons of \emph{CIFAR-only} between these two tables, our 3LSSL method brings significant improvement, \ie, 10.39\%, 5.79\%, 4.89\%, on \emph{MNIST-CIFAR-LT} with different imbalance ratios of $100$, $10$ and $1$, respectively. It is apparent to observe that as the degree of imbalance increases, the accuracy boost by our method becomes more significant. In particular, for tail data, the improvement is 33.76\%, 11.46\% and 5.81\% for the imbalance ratio of $100$, $10$ and $1$. In addition, our method can also improve the classification accuracy in the case of \emph{MNIST-only} by 0.82\%, 0.44\%, 0.21\% for the imbalance ratio of $100$, $10$ and $1$ on such a high basis of 98\%/99\%. These observations could further justify the effectiveness of our method, especially for the tail data. Also, they show that our method does learn more comprehensive information to alleviate simplicity bias in long-tailed data thanks to its SSL designs. Further qualitative visualization can be found in Section~\ref{sec:activis}.

\begin{table*}[t!]
\centering
\small
\renewcommand\arraystretch{1.1}
\caption{\small Classification accuracy (\%) of our proposed 3LSSL method on the constructed \emph{MNIST-CIFAR-LT} dataset with different imbalance ratios (\ie, 100, 10, and 1).}
\begin{tabular}{c|cccc|cccc|cccc}
\toprule
\multirow{2}{*}{\textsc{Test data}} & \multicolumn{4}{c|}{100}        & \multicolumn{4}{c|}{10}         & \multicolumn{4}{c}{1}          \\
\cline{2-13}
                           & \textsf{Head}  & \textsf{Medium} & \textsf{Tail}  & \textsf{All}   & \textsf{Head}  & \textsf{Medium} & \textsf{Tail}  & \textsf{All}   & \textsf{Head}  & \textsf{Medium} & \textsf{Tail}  & \textsf{All} \bigstrut[t]  \\
\hline
\emph{MNIST-CIFAR-LT}      & 99.86 & 99.67 & 99.24 & 99.60 & 99.92 & 99.90 & 99.88 & 99.90 & 99.97 & 99.94 & 99.97 & 99.96 \bigstrut[t]\\
\emph{MNIST-only}          & 99.18 & 99.03 & 97.60 & 98.65 & 99.52 & 99.47 & 99.23 & 99.41 & 99.66 & 99.79 & 99.40 & 99.63 \\
\emph{CIFAR-only}          & 53.82 & 66.21 & 65.92 & 62.41 & 71.94 & 70.38 & 79.62 & 73.62 & 78.95 & 73.44 & 90.27 & 80.14 \\
\bottomrule                
\end{tabular}
\label{table:MNCILT}
\end{table*}

\subsection{Ablation Studies and Discussions}

In this section, we analyze and discuss our proposed 3LSSL by conducting ablation studies on the \emph{ImageNet-LT} dataset.

\subsubsection{Effects of Triple Levels of SSL}

For the crucial components, \ie, triple levels of SSL, in our method, we present the ablation studies of classification accuracy in Table~\ref{table:crucialmodule}, where different settings are created by adding these triple levels of SSL on top of the baseline. More concretely, comparing \#2 with \#1, we can find that the holistic-level SSL improves the overall accuracy by 3.3\%. While, only equipping the partial-level SSL with the baseline obtains 2.7\% improvement, \ie, \#3 vs. \#1. But, when performing both the holistic- and partial-level SSL, it brings 4.5\% accuracy boost (cf. \#4 vs. \#1). These results demonstrate that the holistic-level SSL is fundamental in our 3LSSL, and the partial-level SSL can contribute well as a counterpart w.r.t. the holistic-level SSL. Furthermore, by additionally equipping with the augmented-level SSL, further improvement of 2.1\% is achieved, cf. \#5 vs. \#4. More importantly, the improvement of 2.6\% on tail data on \emph{ImageNet-LT} is obtained, which justifies the effectiveness of the augmented-level SSL w.r.t. tail data.

\begin{table*}[t]
\centering
\small
\renewcommand\arraystretch{1.1}
\caption{\small Ablation studies of classification accuracy (\%) on \emph{ImageNet-LT} w.r.t. the triple levels SSL of our method. Note that, ``\textsf{H-SSL}'', ``\textsf{P-SSL}'' and ``\textsf{A-SSL}'' represent ``Holistic-level SSL'', ``Partial-level SSL'', and ``Augmented-level SSL'', respectively. The highest accuacy is marked in bold.}
\begin{tabular}{c|cccc|cccc}
\toprule
\multirow{2}{*}{\textsc{Number}} & \multicolumn{4}{c|}{\textsc{Settings}}                                     & \multicolumn{4}{c}{\textsc{ImageNet-LT}}                      \\
                                 & \textsf{Baseline (CE)} & \textsf{H-SSL} & \textsf{P-SSL} & \textsf{A-SSL} & \textsf{Many} & \textsf{Med.} & \textsf{Few}  & \textsf{All}  \\
\hline
\#1                              & \checkmark             &                &                &                &        62.7       &      49.3        &        31.6       &   52.5       \bigstrut[t]     \\
\#2                              & \checkmark             & \checkmark     &                &                & 65.9          & 54.3          & 34.2          & 55.8          \\
\#3                              & \checkmark             &                & \checkmark     &                & 64.8          & 53.7          & 34.4          & 55.2          \\
\#4                              & \checkmark             & \checkmark     & \checkmark     &                & 66.8          & 55.4          & 35.7          & 57.0          \\
\#5                              & \checkmark             & \checkmark     & \checkmark     & \checkmark     & \textbf{68.5} & \textbf{57.6} & \textbf{38.3} & \textbf{59.1} \\

\bottomrule           
\end{tabular}
\label{table:crucialmodule}
\end{table*}

\subsubsection{Augmented Queue in the Augmented-level SSL}\label{sec:ablqueue}

We hereby conduct different configurations of the augmented queue in the augmented-level SSL. As aforementioned, the augmented queue is designed to enhance the similarity-based representation learning of tail classes by borrowing pseudo samples from other classes. We consider three configurations, including augmented 1) with a single example $x'$ of class $c'$ in the queue, 2) with all examples of class $c'$ in the queue (\ie, our proposal), and 3) with all examples of class $c$ (\ie, the ground truth class of $x$).

The corresponding results on the \emph{ImageNet-LT} dataset are reported in Table~\ref{table:queue}. Apparently, augmented with a single example will harm the classification accuracy due to the sample randomization and lacking of generalization. More interesting observations can be found by comparing ``2)'' with ``3)''. Compared with our proposal in the augmented queue, augmented with examples belonging to the ground truth obtains 0.2\% overall accuracy drop. In particular, for tail data, a larger drop of 0.5\% is observed, which can clearly demonstrate our effectiveness on tail data by providing chances to see more semantic-relevant samples for learning more robust and comprehensive representations via our augmented-level SSL.

\begin{table}[t]
\centering
\small
\renewcommand\arraystretch{1.1}
\setlength{\tabcolsep}{4.5pt}
\caption{\small Ablation studies of classification accuracy (\%) on \emph{ImageNet-LT} w.r.t. the augmented mechanism in the proposed augmented-level SSL of our method. The highest accuracy is marked in bold. Note that, ``with examples of class $c'$'' is our proposal, which represents that it performs augmented-level SSL on the examples from the predicted label $c'$, rather than those from ground truth label $c$.}
\begin{tabular}{c|cccc}
\toprule
\multirow{2}{*}{\textsc{Setting}} & \multicolumn{4}{c}{\textsc{ImageNet-LT}}                    \\
                                  & \textsf{Many} & \textsf{Med.} & \textsf{Few} & \textsf{All} \\
\hline
With a single example of class $c'$               & 67.4        & 56.3          & 36.9         & 57.8    \bigstrut[t] \\
With examples of class $c$        & \textbf{68.6}           & 57.4          & 37.8         & 58.9         \\
With examples of class $c'$       & 68.5          & \textbf{57.6}          & \textbf{38.3}         & \textbf{59.1}        \\
\bottomrule           
\end{tabular}
\label{table:queue}
\end{table}

\subsubsection{Multi-stage Partial-level SSL}

For the main results of our 3LSSL method, we employ the partial-level SSL by performing only one stage to mask out the class-specific image regions. In the ablation studies of this section, multiple masking stages of the partial-level SSL are conducted, and we report the results in Table~\ref{table:multistageSSL}. As seen, as the number of stages increases, the classification accuracy shows a slight downward trend. The reason might be that with the progress of the masking operation in the partial-level SSL, more and more informative regions are focused and then discarded, and the information in the image regions that are paid attention to in the later stages gradually decreases. Even though our method can force the model to learn comprehensive information from the remaining image regions, the rest information content of these regions is less and less or even non-existent, which could cause redundant and accuracy drop. Therefore, performing one-stage partial-level SSL is both effective and efficient.

\begin{table}[t]
\centering
\small
\renewcommand\arraystretch{1.1}
\caption{\small Ablation studies of classification accuracy (\%) on \emph{ImageNet-LT} w.r.t. the number of stages in the partial-level SSL of our method. The highest accuracy is marked in bold.}
\begin{tabular}{c|cccc}
\toprule
\multirow{2}{*}{\textsc{\# stages}} & \multicolumn{4}{c}{\textsc{ImageNet-LT}}                    \\
                                    & \textsf{Many} & \textsf{Med.} & \textsf{Few} & \textsf{All} \\
\hline
1                                   & \textbf{68.5}          & \textbf{57.6}          & 38.3         & \textbf{59.1}      \bigstrut[t] \\
2                                   & 68.2          & 57.5          & \textbf{38.6}         & 59.0         \\
3                                   & 68.0          & 57.2          & 38.3         & 58.7        \\
\bottomrule           
\end{tabular}
\label{table:multistageSSL}
\end{table}

\subsection{Visualization Analyses}

To further analyze our 3LSSL method, we provide qualitative visualization analyses from the following aspects, \ie, showing feature embedding visualization by $t$-SNE, demonstrating the distribution of classes in the augmented queue, as well as presenting the activation visualization of the learned models.

\subsubsection{Feature Embedding Visualization}

\begin{figure}[t!]
	\centering
	\subfloat[\small {Baseline on training}]  {\includegraphics[width=0.48\columnwidth]{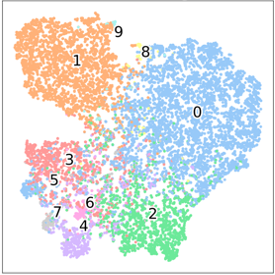} \label{fig:baselineTr}}
	\subfloat[\small {Baseline on test}]  {\includegraphics[width=0.48\columnwidth]{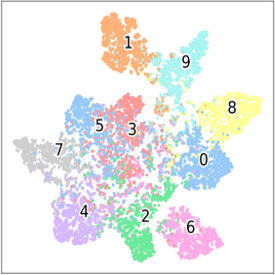} \label{fig:baselineTe}}\\
	\subfloat[\small {Our 3LSSL on training}]  {\includegraphics[width=0.48\columnwidth]{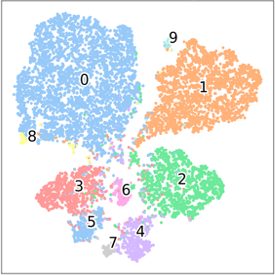} \label{fig:ourTr}}
	\subfloat[\small {Our 3LSSL on test}]  {\includegraphics[width=0.48\columnwidth]{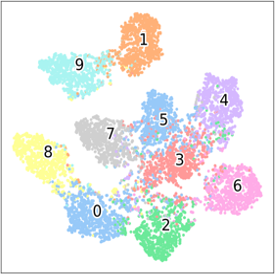} \label{fig:ourTe}}
	\caption{\small $t$-SNE~\cite{tsne} visualization of training and test on the \emph{long-tailed CIFARI-10} dataset (imbalance factor as $100$). The numbers in the figures are class indexes. (Best viewed in color.)}
	\label{fig:tsne}
\end{figure}

In order to intuitively understand the effects of representation learning in 3LSSL, we show $t$-SNE visualization of the learned feature embedding~\cite{tsne} in Figure~\ref{fig:tsne}. We compare the feature embedding learned by the baseline method (\ie, a vanilla backbone) and our 3LSSL of both training and test sets. As shown, the decision boundaries and class separations of the baseline on training and test are all greatly altered by the head classes. While, regardless of the training set or test set, our 3LSSL is able to produce more compact feature embedding and clear class separations, which leads to better discriminative representation learning, especially between adjacent head and tail classes.

\subsubsection{Distribution of Classes in the Augmented Queue}

\begin{figure*}[t!]
	\centering
	\subfloat[\small {\emph{Long-tailed CIFAR-100}}]  {\includegraphics[width=0.5\columnwidth]{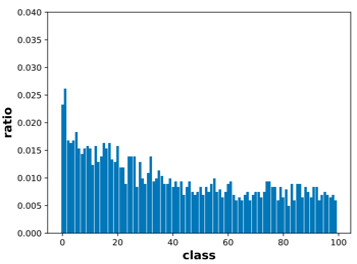} \label{fig:queueCIFAR}}
	\subfloat[\small {\emph{ImageNet-LT}}]  {\includegraphics[width=0.5\columnwidth]{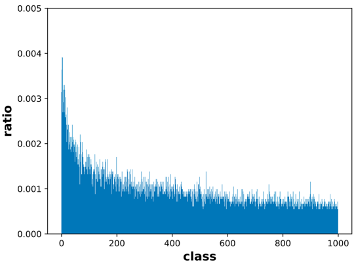} \label{fig:queueImageNet}}
	\subfloat[\small {\emph{Places-LT}}]  {\includegraphics[width=0.5\columnwidth]{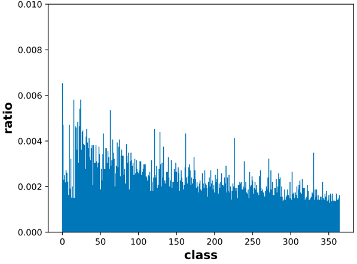} \label{fig:queuePlaces}}
	\subfloat[\small {\emph{iNaturalist 2018}}]  {\includegraphics[width=0.5\columnwidth]{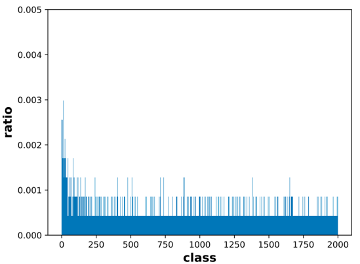} \label{fig:queueiNat}}
	\caption{\small Distribution of classes in the augmented queue on different datasets. We show the distribution after models convengence. The horizontal axis is the class indexes from the head to tail. The vertical axis represents the proportion of the corresponding category to the total samples. For \emph{long-tailed CIFAR-100}, its imbalance factor is $100$. (Best viewed when zoomed in.)}
	\label{fig:queue}
\end{figure*}

\begin{figure*}[p]
	\centering
	\subfloat[\small {{Activations on head data of \emph{MNIST-CIFAR-LT}}}]  {\includegraphics[width=0.7\textwidth]{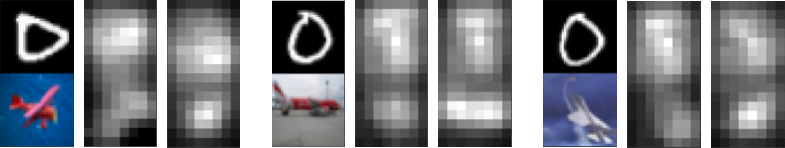} \label{fig:actMCH}}\quad
	\subfloat[\small {Activations on medium data of \emph{MNIST-CIFAR-LT}}]  {\includegraphics[width=0.7\textwidth]{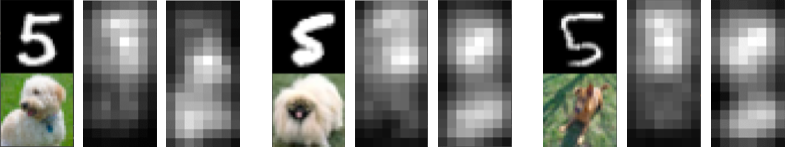} \label{fig:actMCM}}\quad
	\subfloat[\small {Activations on tail data of \emph{MNIST-CIFAR-LT}}]  {\includegraphics[width=0.7\textwidth]{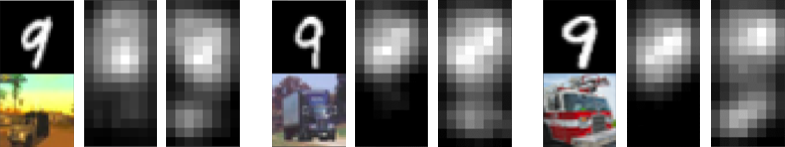} \label{fig:actMCT}}\\
	
	\subfloat[\small {Activations on head data of \emph{ImageNet-LT}}]  {\includegraphics[width=0.7\textwidth]{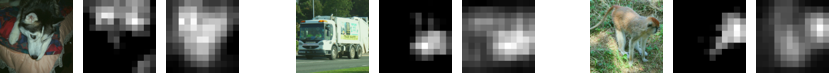} \label{fig:actIMH}}\quad
	\subfloat[\small {Activations on medium data of \emph{ImageNet-LT}}]  {\includegraphics[width=0.7\textwidth]{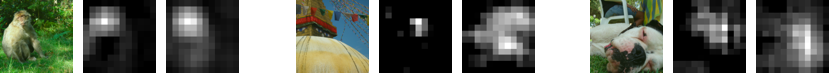} \label{fig:actIMM}}\quad
	\subfloat[\small {Activations on tail data of \emph{ImageNet-LT}}]  {\includegraphics[width=0.7\textwidth]{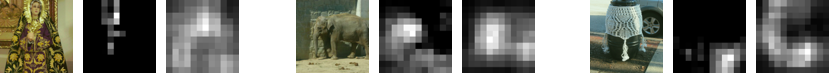} \label{fig:actIMT}}\\
	
	\subfloat[\small {Activations on head data of \emph{Places-LT}}]  {\includegraphics[width=0.7\textwidth]{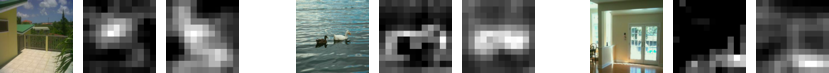} \label{fig:actPLH}}\quad
	\subfloat[\small {Activations on medium data of \emph{Places-LT}}]  {\includegraphics[width=0.7\textwidth]{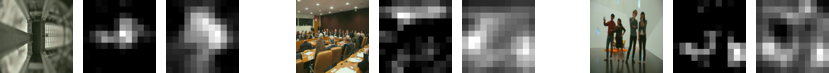} \label{fig:actPLM}}\quad
	\subfloat[\small {Activations on tail data of \emph{Places-LT}}]  {\includegraphics[width=0.7\textwidth]{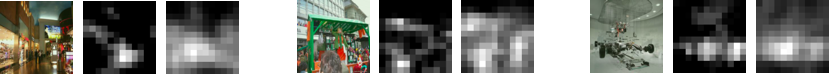} \label{fig:actPLT}}\\
	
	\subfloat[\small {Activations on head data of \emph{iNaturalist 2018}}]  {\includegraphics[width=0.7\textwidth]{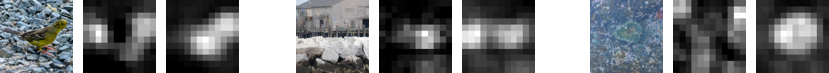} \label{fig:actiNH}}\quad
	\subfloat[\small {Activations on medium data of \emph{iNaturalist 2018}}]  {\includegraphics[width=0.7\textwidth]{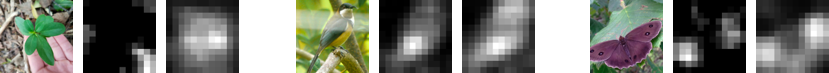} \label{fig:actiNM}}\quad
	\subfloat[\small {Activations on tail data of \emph{iNaturalist 2018}}]  {\includegraphics[width=0.7\textwidth]{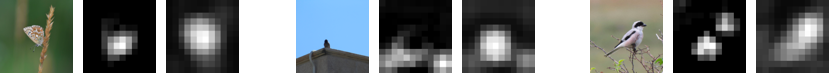} \label{fig:actiNT}}
	\caption{\small Visualization of activations on four datasets. In each sub-figure, the first column is the original image. The second/third column is the activation of training with a vanilla backbone/our 3LSSL method, respectively.}
	\label{fig:activis}
\end{figure*}

As aforementioned in Section~\ref{sec:ablqueue}, we demonstrate the effectiveness of the proposed augmented queue. Hereby, we present the distribution of classes in such an augmented queue for in-depth analyses. After the models are convergence during training, we collect the corresponding augmented queue and show the class distribution in Figure~\ref{fig:queue}. Note that, since \emph{iNaturalist 2018} contains too many categories (\ie, 8,142 fine-grained classes), not all can be presented in the figure. Thus, we uniformly sample 2,000 categories for illustration. As shown in Figure~\ref{fig:queue}, it can be found that compared to their original distribution (being severe long-tailed), the class distribution in the augmented queue is relatively more balanced, especially the \emph{iNaturalist 2018} dataset. More importantly, for the tail data, their proportion in the queue significantly increases, which explains why the augmented queue mechanism of our augmented-level SSL favors tail data.

\subsubsection{Activation Visualization}\label{sec:activis}

To explicitly demonstrate the alleviation of simplicity bias of our method, we visualize the activations of the test images from \emph{MNIST-CIFAR-LT} and three real-world datasets, \ie, \emph{ImageNet-LT}, \emph{Places-LT} and \emph{iNaturalist 2018}, in Figure~\ref{fig:activis}. From these activation visualization, it is obvious to see that our 3LSSL can effectively mitigate the simplicity bias in long-tailed recognition by learning more comprehensive patterns covering the whole informative regions, particularly for the tail data. Such an observation is even more pronounced on the \emph{Places-LT} dataset. We know that scene context plays a crucial role in recognizing scenes. From the visualization results, our method can indeed help the model focus on a more comprehensive scene context region, thereby improving the accuracy of scene recognition.

\section{Conclusion}\label{sec:conc}

In this work, we investigated simplicity bias (SB) in long-tailed image recognition and uncovered underrepresented classes suffer more severely from SB. We empirically reported that self-supervised learning (SSL) can be used as a remedy to mitigate SB and the associated adverse effects through learning more comprehensive features. Following this observation, we proposed a simple but effective method consisting triple levels of SSL (termed as 3LSSL) which is explicitly tailored for long-tailed data to learn image representations that generalize better. Experiments on five long-tailed benchmark datasets show the advantage of the proposed method as well as other appealing properties. We expect our work can provide new understanding about long-tailed learning and inspire new explorations towards SB, including solutions to evade SB, especially in tasks with insufficient data such as few-shot learning. 

There still lacks theoretical support why SSL can mitigate SB. In-depth analyses of the nature of representations learned from SSL is also needed. We will leave such problems in our future work.

\bibliographystyle{IEEEtran}
\bibliography{LTSB}

\begin{IEEEbiography}[{\includegraphics[width=1in,height=1.25in,clip,keepaspectratio]{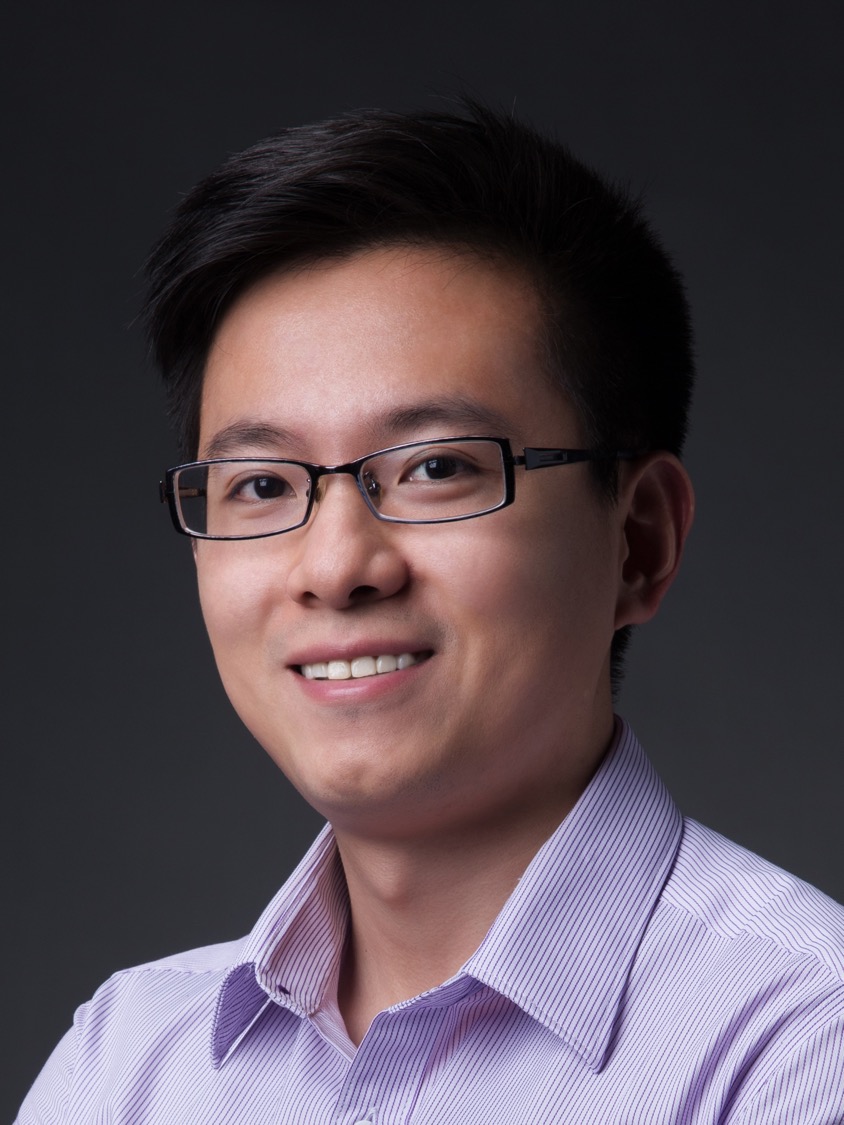}}]{Xiu-Shen Wei} is a Professor with the School of Computer Science and Engineering, Nanjing University of Science and Technology, China. He was a Program Chair for the workshops associated with ICCV, IJCAI, ACM Multimedia, etc. He has also served as an Area Chair or Senior Program Member at AAAI, IJCAI, ICME, BMVC, a Guest Editor of Pattern Recognition Journal, and a Tutorial Chair for Asian Conference on Computer Vision (ACCV) 2022.
\end{IEEEbiography}

\begin{IEEEbiography}[{\includegraphics[width=1in,height=1.25in,clip,keepaspectratio]{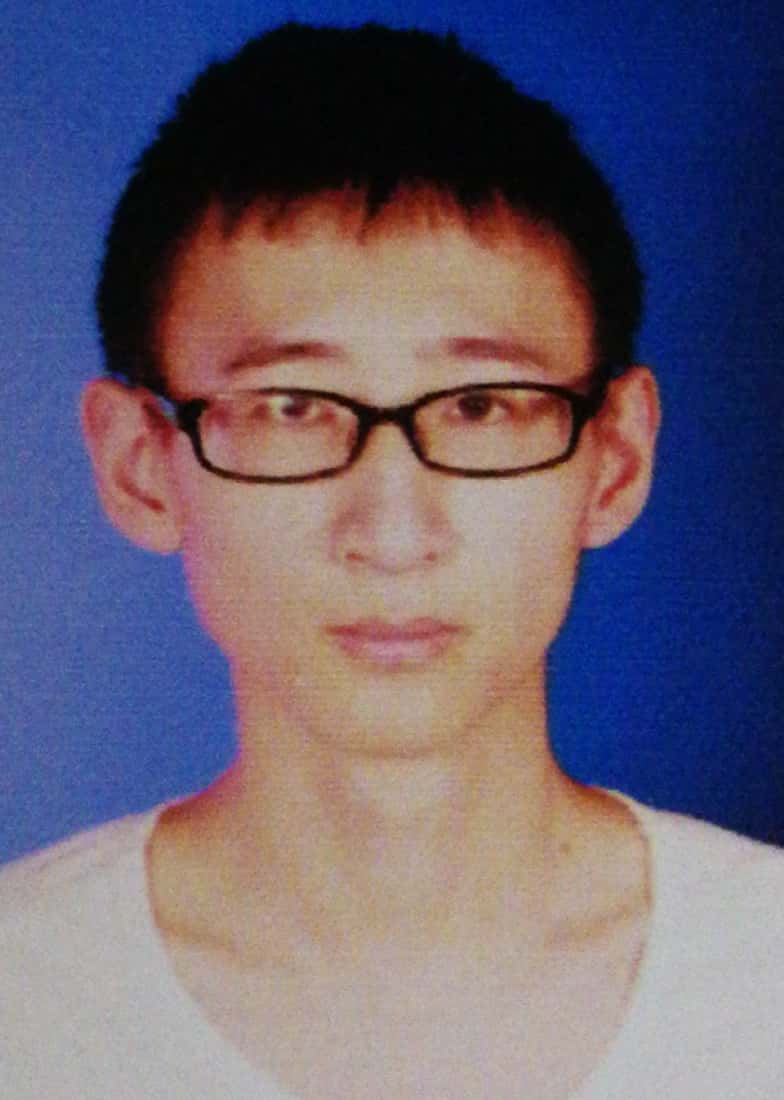}}]{Xuhao Sun} is a Master Student with School of Computer Science and Engineering, Nanjing University of Science and Technology, China. His main research interests are Computer Vision and Machine Learning.
\end{IEEEbiography}

\begin{IEEEbiography}[{\includegraphics[width=1in,height=1.25in,clip,keepaspectratio]{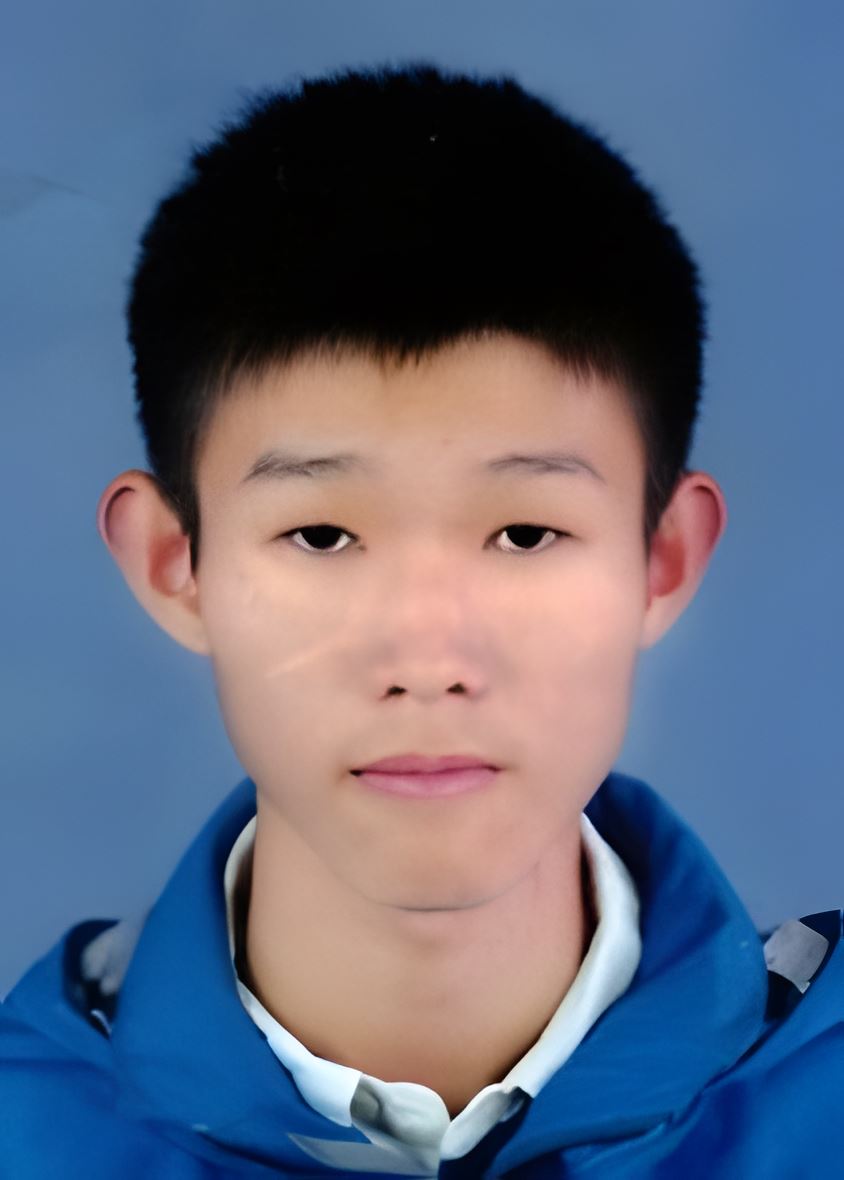}}]{Yang Shen} is a Ph.D Candidate under the supervision of Prof. Xiu-Shen Wei with School of Computer Science and Engineering, Nanjing University of Science and Technology, China. His research interests lie in deep learning and computer vision.
\end{IEEEbiography}

\begin{IEEEbiography}[{\includegraphics[width=1in,height=1.25in,clip,keepaspectratio]{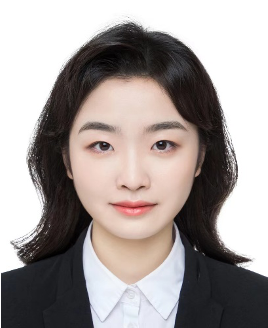}}]{Anqi Xu} is currently studying in Mathematical and Physical Sciences at the University of Toronto, majoring in Statistics and Economics. Anqi graduated from the Shanghai World Foreign Language Academy in which she participated in Yau Science Awards (Economic and Financial) and her paper was accepted by the IEEE-CS (Manuscript title: Industrial Structure Factor Analysis of Complex Network Basis).
\end{IEEEbiography}

\begin{IEEEbiography}[{\includegraphics[width=1in,height=1.25in,clip,keepaspectratio]{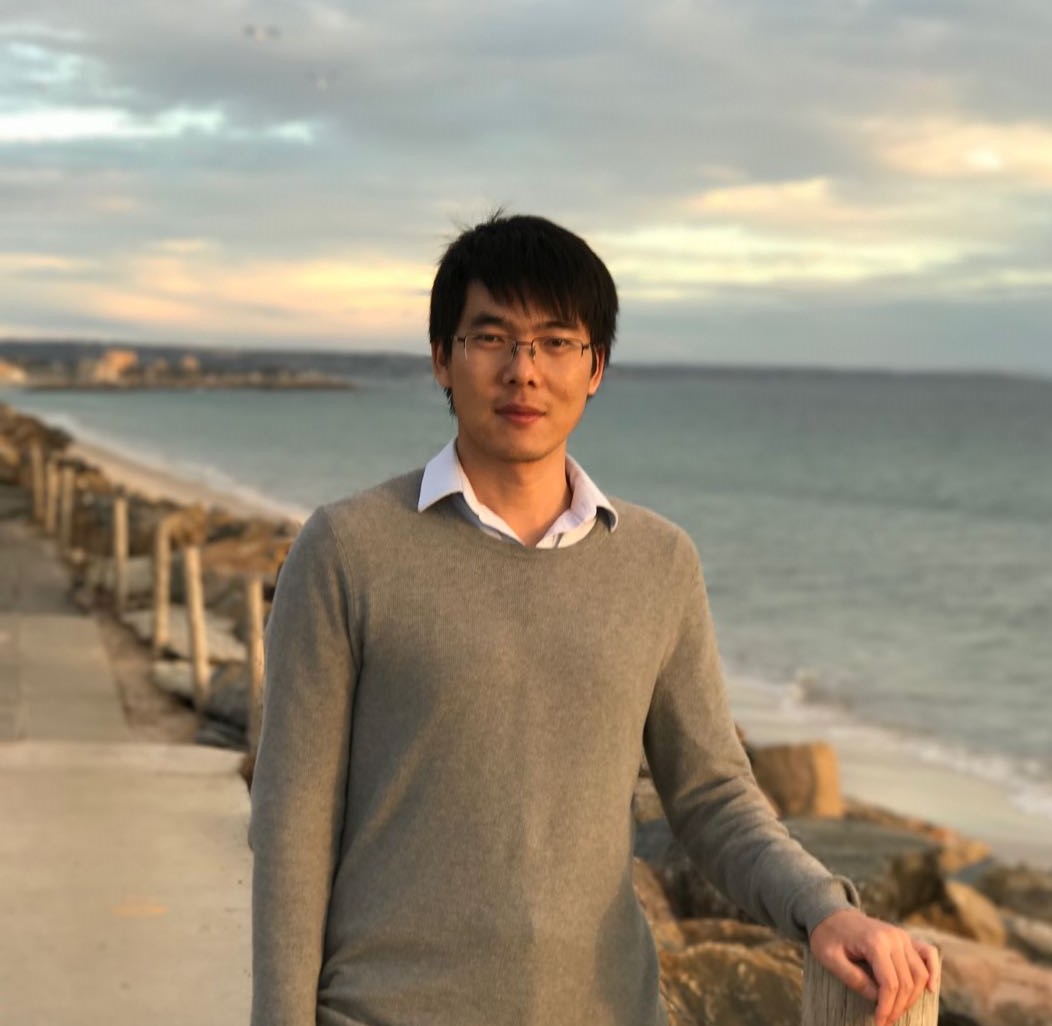}}]{Peng Wang} is a lecturer (assistant professor) with School of Computing and Information Technology at University of Wollongong. Prior to joining UOW, he was a research fellow with Australian Institute for Machine Learning (AIML), The University of Adelaide. He obtained his PhD from School of Information Technology and Electrical Engineering, The University of Queensland. His major research interest lies in computer vision and deep learning, with special interest in data-efficient deep learning.
\end{IEEEbiography}

\begin{IEEEbiography}[{\includegraphics[width=1in,height=1.25in,clip,keepaspectratio]{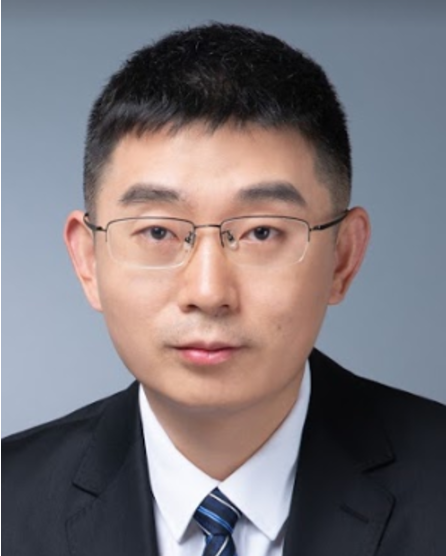}}]{Faen Zhang} received the bachelor's degree in software engineering from Jilin University and the master's degree in computer science and theory from Institute of Software, Chinese Academy of Sciences. He is currently the CTO of AInnovation, the Chief Architect of the AI Institute at Sinovation Ventures and Honorary Professor of Ningbo Nottingham University. He holds 10+ US and 30+ Chinese patents. He has more than 15 years of experience in technology development and management in the IT industry, including enterprise-level software, Internet search engine, knowledge-based maps, big data analysis and storage, machine learning, deep learning. 

\end{IEEEbiography}

\end{document}